\pdfoutput=1

\documentclass[11pt]{article}

\usepackage{EMNLP2023}

\usepackage{times}
\usepackage{latexsym}

\usepackage[T1]{fontenc}

\usepackage[utf8]{inputenc}

\usepackage{microtype}

\usepackage{inconsolata}

\usepackage{graphicx}
\usepackage{amsmath} 
\usepackage{multirow} 
\usepackage{todonotes}
\usepackage{rotating}
\usepackage{booktabs} 
\usepackage{changepage}
\usepackage{algorithmic}
\usepackage{algorithm}
\usepackage{xcolor}
\usepackage{amsfonts}

\title{MiniConGTS: A Near Ultimate Minimalist Contrastive Grid Tagging Scheme for Aspect Sentiment Triplet Extraction}

\author{Qiao Sun$^{1,2}$ \quad Liujia Yang$^{1,4}$ \quad Minghao Ma$^{1,3}$ \quad Nanyang Ye$^{4*}$ \quad Qinying Gu$^{1*}$ \\
$^{1}$Shanghai Artificial Intelligence Laboratory, China \\
$^{2}$Academy of Engineering and Technology, Fudan University, China \\
$^{3}$School of Information Science and Technology, Fudan University, China \\
$^{4}$Shanghai Jiao Tong University, China \\
\texttt{qiaosun22@m.fudan.edu.cn} \quad \texttt{20307130024@fudan.edu.cn} \\
\texttt{yangliujia1008@sjtu.edu.cn} \quad \texttt{ynylincolncam@gmail.com} \quad \texttt{guqinying@pjlab.org.cn} \\
}
\begin{document}
\maketitle
\renewcommand{\thefootnote}{\fnsymbol{footnote}}
\footnotetext[1]{Corresponding Authors}



\maketitle
\begin{abstract}
Aspect Sentiment Triplet Extraction (ASTE) aims to co-extract the sentiment triplets in a given corpus. 
Existing approaches within the pretraining-finetuning paradigm tend to either meticulously craft complex tagging schemes and classification heads, or incorporate external semantic augmentation to enhance performance. 
In this study, we, for the first time, re-evaluate the redundancy in tagging schemes and the internal enhancement in pretrained representations.  
We propose a method to improve and utilize pretrained representations by integrating a minimalist tagging scheme and a novel token-level contrastive learning strategy. 
The proposed approach demonstrates comparable or superior performance compared to state-of-the-art techniques while featuring a more compact design and reduced computational overhead. 
Additionally, we are the first to formally evaluate GPT-4's performance in few-shot learning and Chain-of-Thought scenarios for this task. The results demonstrate that the pretraining-finetuning paradigm remains highly effective even in the era of large language models. The codebase is available at \href{https://github.com/qiaosun22/MiniConGTS}{https://github.com/qiaosun22/MiniConGTS}.
\end{abstract}

\section{Introduction}
\begin{figure}[ht]
    \centering
    \includegraphics[width=1\linewidth]{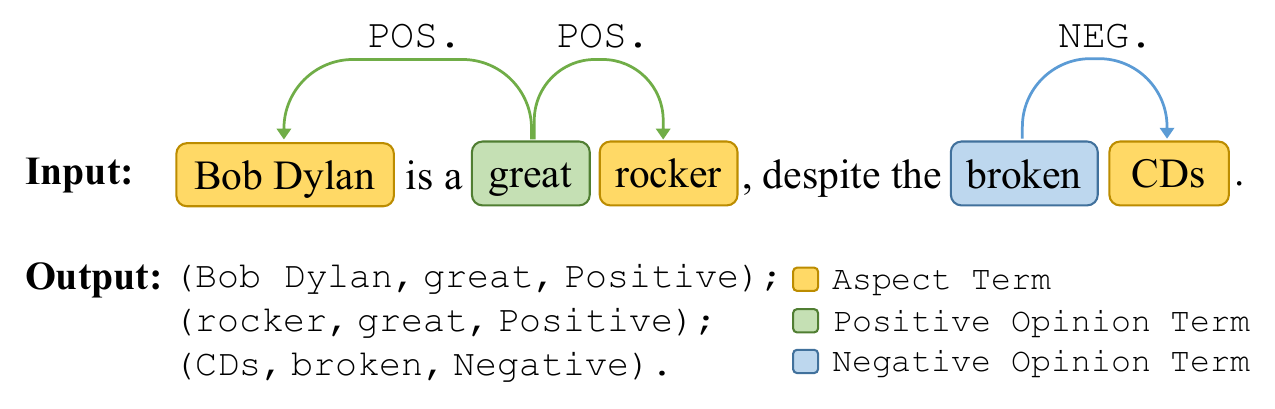}
    \caption{An illustration for ASTE, given the sentence "Bob Dylan is a great rocker, despite the broken CDs.", there are three triplets to be extracted: \texttt{(Bob Dylan, great, positive)}, \texttt{(rocker, great, positive)}, \texttt{(CDs, broken, negative)}. }
    \label{fig:fig_task.pdf}
\end{figure}

Aspect-Based Sentiment Analysis (ABSA) aims to jointly extract opinion terms, aspect terms (targets of the corresponding opinions), and their specific sentiment polarities in a given corpus. In the milestone research by \citeauthor{peng2020knowing} (\citeyear{peng2020knowing}), the compound ABSA subtasks were consolidated into the Aspect Sentiment Triplet Extraction (ASTE) task framework. For each input corpus, ASTE outputs triplets in the form \texttt{(Aspect, Opinion, Polarity)}, where the \texttt{Aspect} term is the target or entity being discussed, the \texttt{Opinion} term is the sentiment or opinion expressed about the aspect, and \texttt{Polarity} indicates whether the opinion is positive, negative, or neutral. Figure \ref{fig:fig_task.pdf} illustrates the ASTE task.

As an emerging fine-grained sentiment analysis initiative, ASTE offers a more detailed and nuanced understanding of sentiments in text compared to traditional methods that provide only an overall sentiment score \cite{peng2020knowing}. This aspect-level structured approach is inherently more challenging.

Previous approaches to ASTE have generally followed two paradigms: Pipeline methods and Joint Tagging methods \cite{zhang2022survey}. 
Pipeline methods decompose the ASTE task into multiple sequential subtasks, often suffering from error propagation \cite{xu2020position}. Recent progress in Machine Reading Comprehension (MRC) also contributes to this paradigm \cite{zhai2022mrc, mao2021joint, zou2024multi, chen2021semantic}. 
Joint tagging methods adopt a unified tagging scheme to extract all triplet elements in one stage \cite{xu2020position}. The key idea is to design a \textbf{tagging scheme} \cite{zheng2017joint} that simultaneously predicts aspect terms, opinion terms, and sentiment polarities. Further developments have introduced a Grid Tagging Scheme (GTS) to represent the triplets on a unique 2D table\cite{wu2020grid,zhang2022boundary,chen2021semantic,chen2022enhanced,fei2022inheriting}.

Recent advances in these approaches have been focusing on the classification head design \cite{chen2022enhanced, zhang2022boundary} and external semantic information enhancing \cite{chen2021semantic, chen2022enhanced, fei2022inheriting, jiang2023semantically, iswari2024enhancing}. However, existing research has neglected the synergistic optimization within the joint tagging scheme and the integration of contextual word representations. In this research, we propose a method to effectively improve and utilize the representation capabilities of pretrained encoders in ABSA by integrating a minimalist tagging scheme and a novel token-level contrastive learning approach. 

The proposed approach demonstrates comparable or superior performance in comparison to state-of-the-art techniques, while featuring a more compact design and reduced computational overhead. Notably, even in the era of Large Language Models (LLMs), our method exhibits superior effectiveness compared to GPT 3.5 and GPT 4 in both few-shot and Chain-of-Thought \cite{wei2022chain} learning scenarios. This study provides valuable insights for the advancement of ASTE techniques within the paradigm of LLMs. Overall, our contributions are summarized as follows:

\vspace{-2mm}
\begin{enumerate}
    \item \textbf{Minimalist Grid Tagging Scheme}: We propose a novel minimalist joint tagging scheme that uses the fewest label classes to date.
    \vspace{-3mm}
    \item \textbf{Token-level Contrastive Learning Strategy}: We introduce a token-level contrastive learning framework that enhances the contextual embeddings produced by the pretrained model. This framework is seamlessly geared towards our minimalist Grid Tagging Scheme (GTS) to effectively address the ASTE task. 
    \vspace{-3mm}
    \item \textbf{Comprehensive Evaluation}: 
    We conduct extensive experiments and evaluations on multiple benchmark datasets, demonstrating the effectiveness and superiority of our proposed methods over existing approaches. Notably, we are the first to reveal GPT-4's performance on this task, showcasing our method’s superior efficiency and effectiveness in the era of large language models.
\end{enumerate}

\section{Literature Review}
\subsection{ASTE Paradigms}
\citeauthor{peng2020knowing} (\citeyear{peng2020knowing}) proposed a pipeline method that divides ASTE tasks into two stages: initially extracting \texttt{(Aspect, Opinion)} pairs and subsequently predicting sentiment polarity. However, pipeline methods typically suffer from error propagation issues \cite{xu2020position}. Recent pipeline methods treat ASTE as a Machine Reading Comprehension problem, and develops seq2seq methods such as machine reading comprehension \cite{zhai2022mrc, mao2021joint, zou2024multi, chen2021semantic}. 
Joint Tagging strategies are remarked by certain Unified Tagging Scheme designs, where elements of a triplet can be extracted simultaneously. ET \cite{xu2020position} introduced a position-aware tagging scheme with a conditional random field module, effectively addressing span overlapping issues.
Recent joint paradigm methods have refined the ASTE task with the development of proficient Grid Tagging Schemes (GTS).

\subsection{Grid Tagging Scheme}
\citeauthor{wu2020grid} (\citeyear{wu2020grid}) pioneered the adoption of a grid tagging scheme (GTS) for ASTE, yielding substantial performance gains. 
Subsequent research refined and enhanced GTS. 
BDTF \cite{zhang2022boundary} designed a boundary-driven tagging scheme, effectively reducing boundary prediction errors. 
Alternative research augmented GTS by integrating external semantic information as structured knowledge into their models. $\mathrm{S}^3\mathrm{E}^2$ \cite{chen2021semantic} retained the GTS tagging scheme while introducing novel semantic and syntactic enhancement modules between word embedding outputs and the tagging scheme. EMGCN \cite{chen2022enhanced} incorporated external knowledge from four areas—Part-of-Speech Combination, Syntactic Dependency Type, Tree-based Distance, and Relative Position Distance—through an exogenous hard-encoding strategy.
SyMux \cite{fei2022inheriting} contributed a unified tagging scheme capable of handling all ABSA subtasks by integrating insights from GCN, syntax encoders, and representation multiplexing.

\subsection{Contrastive Learning}
While contrastive learning has gained popularity in diverse NLP domains \cite{wu2020clear, giorgi2021declutr, gao2021simcse, zhang2021supporting}, its application to ASTE remains relatively unexplored. \citeauthor{ye2021contrastive} (\citeyear{ye2021contrastive}) adopts contrastive learning into triplet extraction in a generative fashion. \citeauthor{wang2022contrastive} (\citeyear{wang2022contrastive}) takes contrastive learning as a data augmentation approach. \citeauthor{yang2023pairing} (\citeyear{yang2023pairing}) proposed an enhancement approach in pairing with two separate encoders.

\section{Method}
\subsection{Overall Framework}
\label{sec:Overall Framework}
\begin{figure}[ht]
    \centering
    \includegraphics[width=1\linewidth]{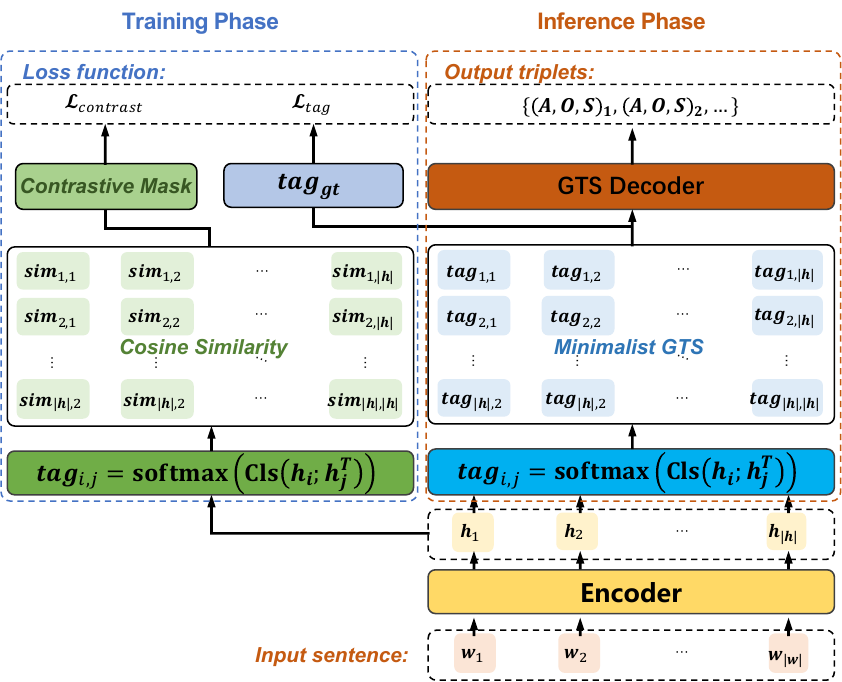}
    \caption{An overview of the proposed method, where the ``Encoder'' denotes for the sequential combination of a Tokenizer and a Pretrained Language Model (PLM). }
    \label{fig:fig_framework.pdf}
\end{figure}

An overall description of the training process can be found in Figure \ref{fig:fig_framework.pdf}.  Basically our design can be break down into the \textbf{Minimalist Grid Tagging Scheme (GTS)} and the \textbf{Token-level Contrastive Learning Strategy}.

Tokenize the input sequence \(\textit{\textbf{S}}\) using the Tokenizer \(\mathrm{Tk}\) and pass the tokenized sequence through the Pretrained Language Model \(\mathrm{PLM}\) (such as BERT) to obtain contextualized representations \(\textit{\textbf{h}}\): 
\begin{equation}
\setlength{\abovedisplayskip}{2pt}
\setlength{\belowdisplayskip}{2pt}
\textit{\textbf{h}} = \mathrm{PLM}\left( \mathrm{Tk}\left( \textit{\textbf{S}} \right) \right).
\label{eq:encoder}
\end{equation}

Then, the inference phase involves with forming the Minimalist GTS and predicting the corresponding class for each cell. Once the GTS is predicted, it can be decoded by the GTS decoder into the triplets in natural language form. The training phase additionally introduces a novel contrastive learning strategy, where similar and dissimilar pairs of contextual representations are distinguished. The contrastive loss is then weighted and summed with the tagging loss, which is the classification loss between the predicted and ground truth tagging schemes.

Our research benefits from the following two closely intertwined aspects: 1) The use of the Minimalist GTS simplifies the learning process by reducing the number of labels, facilitating faster convergence and seamlessly gearing the contrastive learning. 2) The token-level contrastive learning enhances the model's ability to distinguish between related and unrelated elements within the input sequence, thereby improving the overall accuracy of the tagging system.
For a more detailed description for our algorithm pipeline, see the pseudo code in Appendix \ref{sec:pseudo}.

\subsection{Minimalist Grid Tagging Scheme}
\subsubsection{Tagging Scheme Design}

\begin{figure}[ht]
    \centering
    \includegraphics[width=1\linewidth]{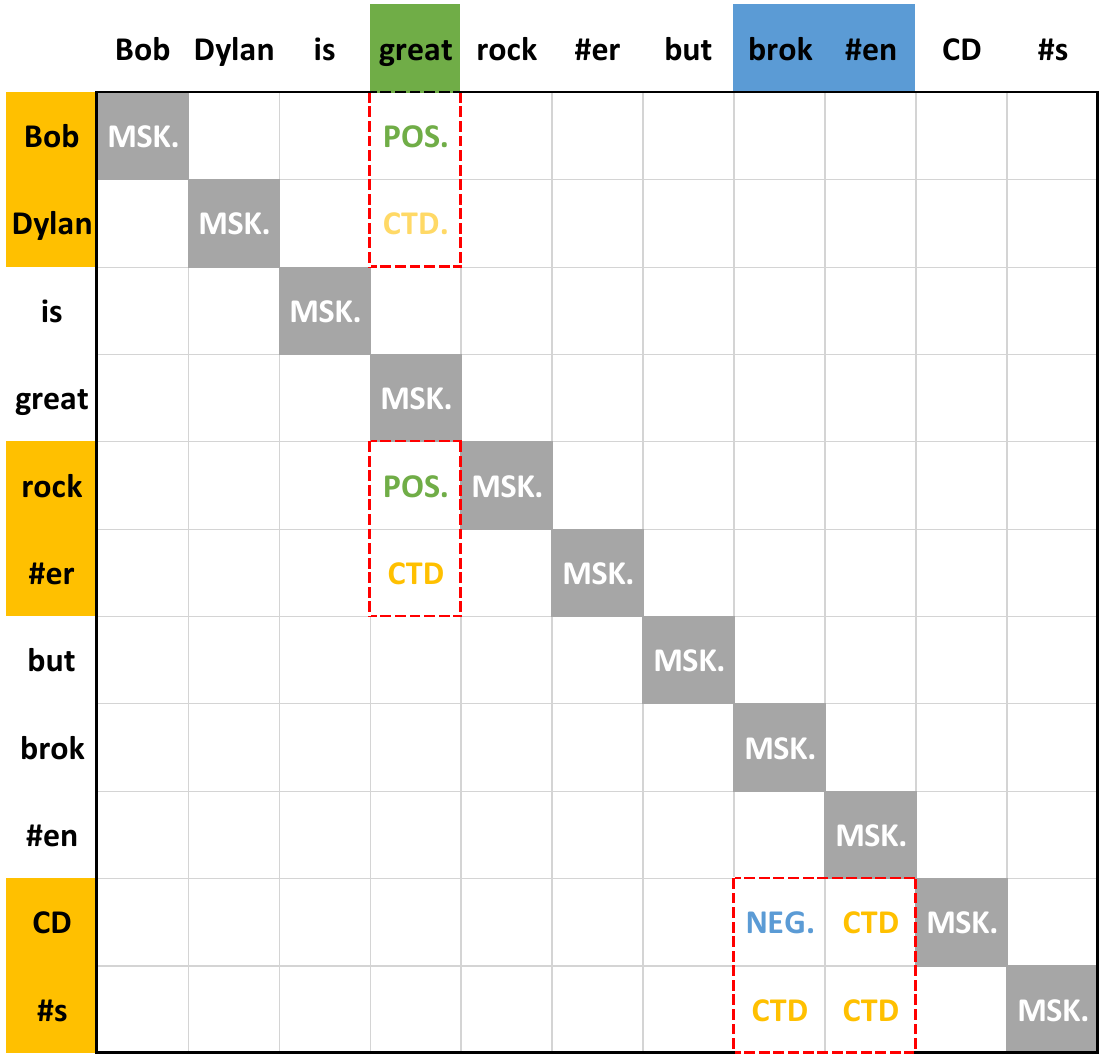}
    \caption{The grid tagging scheme employs the fewest classes of labels while completely handle all the triplet cases without conflict, overlap or omission. Each \textbf{area} circled in \textcolor{red}{red} dashed lines corresponds to a triplets. For example, 
    intersection area between columns of "broken" and rows of "CDs" is marked as negative, with \texttt{NEG.} on its top-left cell and \texttt{CTD.} for others. It is worth mentioning that the blank cells in the matrix are labeled as an additional class but are omitted for visual simplicity.}
    \label{fig_grid.pdf}
\end{figure}

\noindent
As defined by Section \ref{sec:Overall Framework}, once an input sentence is encoded into a sequence of contextual representations \(\textit{\textbf{h}}=\{h_1, h_2, ..., h_{|\textit{\textbf{h}}|}\}\), we form a \(|\textit{\textbf{h}}| \times  |\textit{\textbf{h}}|\) matrix, that is, our tagging scheme \(\textit{\textbf{tag}}_{|\textit{\textbf{h}}| \times  |\textit{\textbf{h}}|} \). 
As shown in Figure \ref{fig_grid.pdf}, on the rows we mark Aspect tokens by \textcolor{yellow}{yellow} and the columns we mark Opinion tokens by \textcolor{green}{green} (positive) and \textcolor{blue}{blue} (negative). Then the each intersection of these marked rows and columns can uniquely represent an identical sentiment triplet. Thus, each such triplet can be noted by a 2-D \textbf{area} (submatirx) in the matrix, where Sentiment Polarity is indicated with \texttt{POS.} (positive), \texttt{NEU.} (neutral), or \texttt{NEG.} (negative) in the top-left corner cell of the area, while \texttt{CTD.} indicates the continuation of the pairing relationship within the same region. \texttt{MSK.} (mask) on the diagonal represents masked cells that are not involved in the computation. In Figure \ref{fig_grid.pdf}, an example sentence is tokenized and tagged. 

By defining our grid tagging scheme, we frame the triplet extraction problem as a 5-class classification task, using the fewest number of labels known to date. In Appendices \ref{Rethinking the GTS} - \ref{Proof 2}, we provide rigorous proof and heuristic insights to justify our design and ensure its rationality.


\subsubsection{Tagging Loss}

We adopt a tagging loss to guide neural network learning.

We concatenate the representation with its transposed form to construct a matrix. Then, we apply the classification head \(\mathrm{Cls}\) to the embeddings, followed by the softmax function, to obtain the predicted classification probabilities for each cell: 
\begin{equation}
   \setlength{\abovedisplayskip}{2pt}
   \setlength{\belowdisplayskip}{2pt}
   \hat{\textit{\textbf{tag}}}_{i,j} = \text{softmax}(\mathrm{Cls}(\textit{\textbf{h}}_i; \textit{\textbf{h}}_j^\mathrm{T})), i, j = 1, ..., |\textit{\textbf{h}}|.
   \label{eq:prediction}
\end{equation}

The focal loss \cite{lin2017focal} is employed to mitigate class imbalance by placing greater emphasis on examples that are difficult to classify correctly. This is achieved by down-weighting the loss for well-classified instances and focusing more on misclassified instances. The formula for focal loss \(\mathcal{L}\) is as follows:
\begin{equation}
\setlength{\abovedisplayskip}{2pt}
\setlength{\belowdisplayskip}{2pt}
    \mathcal{L}_{\text{tag}} = - \frac{1}{|\textit{\textbf{h}}|^2} \sum_{i,j=1}^{|\textit{\textbf{h}}|} \alpha_{\textit{\textbf{tag}}_{i,j}} (1 - \textit{tag}_{t})^\gamma \log(\textit{tag}_{t}), 
\label{eq:focal_loss}
\end{equation}
where \(\alpha\) is a weighting factor for balancing the importance of tags, \(\gamma\) is a focusing parameter that increases the weight of hard-to-predict tags, and \(\textit{\textbf{tag}}_{i,j}\) and \(tag_t\) represent the ground truth label and the predicted probability for the true label at position \((i, j)\), respectively:
\begin{equation}
\setlength{\abovedisplayskip}{2pt}
\setlength{\belowdisplayskip}{2pt}
    tag_t=\hat{\textit{\textbf{tag}}}_{i,j;\textit{\textbf{tag}}_{i,j}}
\label{eq:focal_loss_tag}
\end{equation}

\subsection{Contrastive Learning Strategy}
\subsubsection{Contrastive Learning Label Matrix}
\begin{figure}[ht]
    \centering
    \includegraphics[width=1\linewidth]{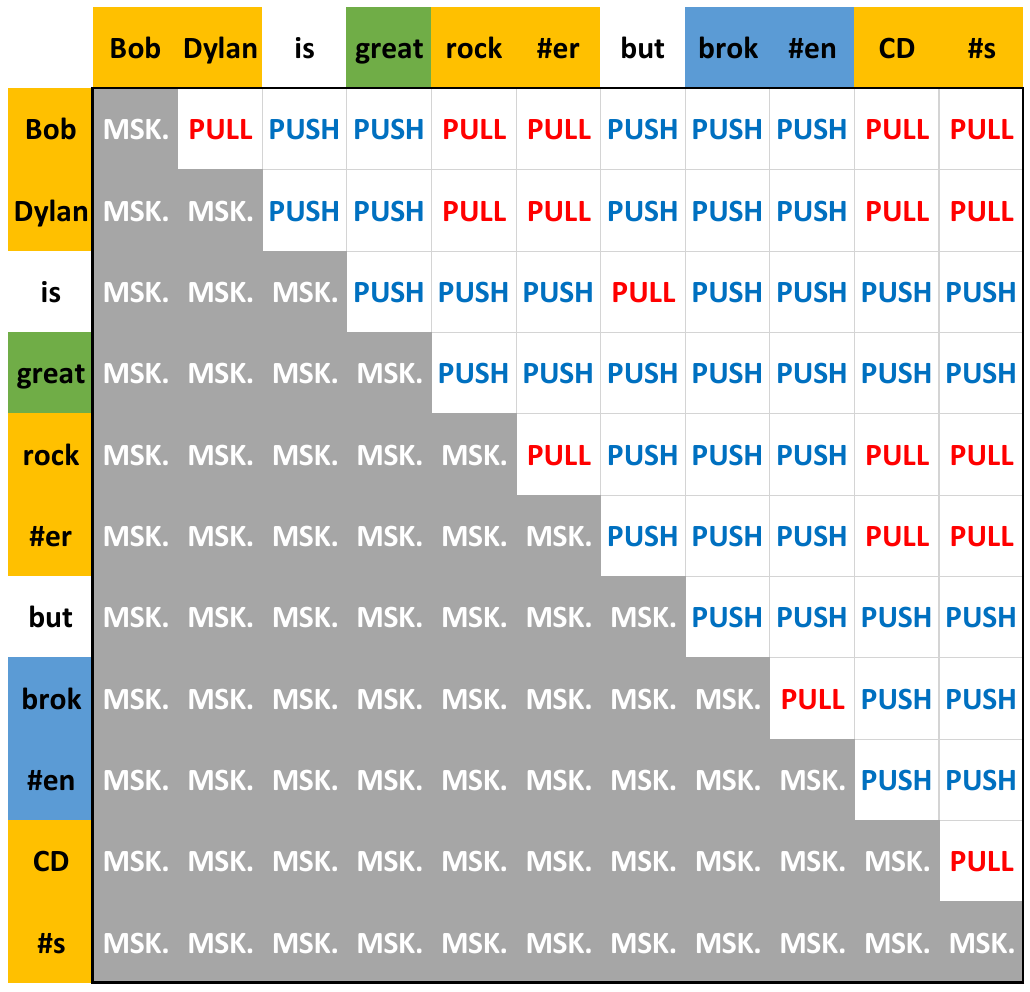}
    \caption{An illustration for the ``Contrastive Mask''. Each token is paired with every other token, where \texttt{PULL} denotes positive sample pairs, indicating that the tokens belong to the same category and should be pulled closer together, while \texttt{PUSH} denotes negative sample pairs, indicating that the tokens belong to different categories and should be pushed apart. The  lower triangular part of the matrix, marked by \texttt{MSK.} are masked cells that are not involved in the computation. For example, "Bob" and "Dylan" are marked as a positive sample pair with \texttt{PULL}, indicating similarity, while "Bob" and "is" are marked as a negative sample pair with \texttt{PUSH}, indicating dissimilarity.}
    \label{fig:fig_contr.pdf}
\end{figure}

Contrastive learning is an unsupervised learning method that aims to learn effective feature embeddings by pulling together similar pairs of samples and pushing apart dissimilar pairs. In our design, we construct a label matrix where each cell is annotated by either PULL or PUSH, which means making the representations closer among tokens within the same class and farther between those of different classes. 
See an illustration of this strategy in Figure \ref{fig:fig_contr.pdf}. 

\subsubsection{Objective Function}

The commonly used InfoNCE (Information Noise-Contrastive Estimation) loss function \cite{oord2019representation} is employed:
\begin{align}
\setlength{\abovedisplayskip}{0pt}
\setlength{\belowdisplayskip}{0pt}
 & \scalebox{1}{$\mathcal{L}_{\text{contrast}}$} \notag \\
=& \ \scalebox{0.8}{$-\displaystyle\sum_{i=1}^{N} \log \frac{\exp(\text{sim}(\textit{\textbf{h}}_i, \textit{\textbf{h}}_i^+))}{\exp(\text{sim}(\textit{\textbf{h}}_i, \textit{\textbf{h}}_i^+)) + \displaystyle\sum_{j=1}^{M} \exp(\text{sim}(\textit{\textbf{h}}_i, \textit{\textbf{h}}_i^-))} $}, 
\end{align}
where $\textit{\textbf{h}}_i^+$ / $\textit{\textbf{h}}_j^-$ represents the positive / negative sample embedding with the anchor respectively. $\text{sim}(\cdot, \cdot)$ denotes the similarity function, which is calculated by the cosine similarity:
\begin{equation}
\setlength{\abovedisplayskip}{2pt}
\setlength{\belowdisplayskip}{2pt}
\text{sim}(\textit{\textbf{h}}_i, \textit{\textbf{h}}_i) = \frac{\textit{\textbf{h}}_i \cdot \textit{\textbf{h}}_i}{\|\textit{\textbf{h}}_i\| \|\textit{\textbf{h}}_i\|}
\end{equation}

\subsection{Overall Loss Function}

The overall loss \( \mathcal{L} \) can be formulated as a weighted sum of two individual loss functions: the tagging loss \( \mathcal{L}_{tag} \) and the contrastive loss \( \mathcal{L}_{contrast} \):
\begin{equation}
\setlength{\abovedisplayskip}{2pt}
\setlength{\belowdisplayskip}{2pt}
    \mathcal{L} = \mathcal{L}_{tag} + \beta \mathcal{L}_{contrast}, 
\label{eq:overall_loss}
\end{equation}
where $\beta$ is a coefficient for balancing the different parts of the loss.
This combination allows for balancing the influence of each component in the training process.

\section{Experiments}


\subsection{Implementation Details}

All experiments were performed on a single RTX 2080 Ti. 
The best model weight on the development set is saved and then evaluated on the test set. For the PLM encoder, the pretrained weights \texttt{bert\_base\_uncased} and \texttt{roberta\_base} are downloaded from \citep{wolf2020transformers}. GPT 3.5-Turbo and GPT 4 are implemented using OpenAI API \cite{openai_api}.
The learning rate is $1\times 10^{-5}$ for the PLM encoder, and $1\times 10^{-3}$ for the classification head. 

\subsection{Datasets}
We evaluate our method on two canonical ASTE datasets derived from the SemEval Challenges \cite{pontiki2014novel, pontiki2015semeval, pontiki2016semeval}. These datasets serve as benchmarks in most aspects-based sentiment analysis (ABSA) research. The first dataset, denoted as $\mathcal{D}_1$, is the Aspect-oriented Fine-grained Opinion Extraction (AFOE) dataset introduced by \cite{wu2020grid}. The second dataset, denoted as $\mathcal{D}_2$, is a refined version by \cite{xu2020position}, building upon the work of \cite{peng2020knowing}. More details are provided in Appendix \ref{sec:Descriptive Statistics of The Datasets}.

\subsection{Baselines}
We evaluate our method against various techniques including pipeline, sequence-labeling, seq2seq, table-filling and LLM-based approaches. Detailed descriptions for each method can be found in the Appendix \ref{sec:Baseline}.

\subsection{ASTE Performance}
\subsubsection{Comparison to Existing Methods}
\begin{table*}[ht]
\centering
  \scalebox{0.68}{
    \begin{tabular}{cccccccccccccccc}
    \hline
    \toprule
    \multirow{2}[4]{*}{\textbf{Methods}} & \multicolumn{3}{c}{\textbf{14Res}} &       & \multicolumn{3}{c}{\textbf{14Lap}} &       & \multicolumn{3}{c}{\textbf{15Res}} &       & \multicolumn{3}{c}{\textbf{16Res}} \\
\cmidrule{2-4}\cmidrule{6-8}\cmidrule{10-12}\cmidrule{14-16}          & P     & R     & F1    &       & P     & R     & F1    &       & P     & R     & F1    &       & P     & R     & F1 \\
    \midrule
    \midrule
    \textbf{Pipeline} &       &       &       &       &       &       &       &       &       &       &       &       &       &       &  \\
    $\textrm{Two-stage}^\natural$\ \textrm{\cite{peng2020knowing}} & 43.24 & 63.66 & 51.46 &       & 37.38 & 50.38 & 42.87 &       & 48.07 & 57.51 & 52.32 &       & 46.96 & 64.24 & 54.21 \\
    $\textrm{Li-unified-R+PD}^\sharp$\ \textrm{\cite{peng2020knowing}} & 40.56 & 44.28 & 42.34 &       & 41.04 & 67.35 & 51.00 &       & 44.72 & 51.39 & 47.82 &       & 37.33 & 54.51 & 44.31 \\
    \midrule
    \textbf{Sequence-tagging} &       &       &       &       &       &       &       &       &       &       &       &       &       &       &  \\
    Span-BART \textrm{\cite{yan2021unified}}  & 65.52 & 64.99 & 65.25 &       & 61.41 & 56.19 & 58.69 &       & 59.14 & 59.38 & 59.26 &       & 66.60 & 68.68 & 67.62 \\
    JET \textrm{\cite{xu2020position}}  & 70.56 & 55.94 & 62.40 &       & 55.39 & 47.33 & 51.04 &       & 64.45 & 51.96 & 57.53 &       & 70.42 & 58.37 & 63.83 \\
    \midrule
    \textbf{Seq2seq} &       &       &       &       &       &       &       &       &       &       &       &       &       &       &  \\
    Dual-MRC \textrm{\cite{mao2021joint}} & 71.55 & 69.14 & 70.32 &       & 57.39 & 53.88 & 55.58 &       & 63.78 & 51.87 & 57.21 &       & 68.60 & 66.24 & 67.40 \\
    $\textrm{BMRC}^\dagger$ \textrm{\cite{chen2021bidirectional}} & 72.17 & 65.43 & 68.64 &       & 65.91 & 52.15 & 58.18 &       & 62.48 & 55.55 & 58.79 &       & 69.87 & 65.68 & 67.35 \\
    COM-MRC \textrm{\cite{zhai2022mrc}} & 75.46 & 68.91 & 72.01 &       & 62.35 & 58.16 & 60.17 &       & 68.35 & 61.24 & 64.53 &       & 71.55 & 71.59 & 71.57 \\ 
    Triple-MRC \textrm{\cite{zou2024multi}} & - & - & 72.45 &       & - & - & 60.72 &       & - & - & 62.86 &       & - & - & 68.65 \\
    
    \midrule
    \textbf{Table-filling} &       &       &       &       &       &       &       &       &       &       &       &       &       &       &  \\
    GTS \textrm{\cite{wu2020grid}}   & 67.76 & 67.29 & 67.50 &       & 57.82 & 51.32 & 54.36 &       & 62.59 & 57.94 & 60.15 &       & 66.08 & 66.91 & 67.93 \\
    Double-encoder \textrm{\cite{jing2021seeking}} & 67.95 & 71.23 & 69.55 &       & 62.12 & \underline{56.38} & 59.11 &       & 58.55 & 60.00 & 59.27 &       & 70.65 & 70.23 & 70.44 \\
    EMC-GCN \textrm{\cite{chen2022enhanced}} & 71.21 & 72.39 & 71.78 &       & 61.70 & 56.26 & 58.81 &       & 61.54 & 62.47 & 61.93 &       & 65.62 & 71.30 & 68.33 \\

    BDTF \textrm{\cite{zhang2022boundary}} & 75.53 & \underline{73.24} & \underline{74.35} &       & 68.94 & 55.97 & 61.74 &       & 68.76 & \underline{63.71} & \textbf{66.12} &       & 71.44 & \underline{73.13} & 72.27 \\
    STAGE-1D \textrm{\cite{liang2023stage}} & \textbf{79.54} & 68.47 & 73.58 &       & \underline{71.48} & 53.97 & 61.49 &       & 72.05 & 58.23 & 64.37 &       & \textbf{78.38} & 69.10  & \underline{73.45} \\
    STAGE-2D \textrm{\cite{liang2023stage}} & 78.51 & 69.3  & 73.61 &       & 70.56 & 55.16 & \underline{61.88} &       & \underline{72.33} & 58.93 & 64.94 &       & \underline{77.67} & 68.44 & 72.75 \\
    STAGE-3D \textrm{\cite{liang2023stage}} & \underline{78.58} & 69.58 & 73.76 &       & \textbf{71.98} & 53.86 & 61.58 &       & \textbf{73.63} & 57.9  & 64.79 &       & 76.67 & 70.12 & 73.24 \\
    DGCNAP \textrm{\cite{li2023dual}} & 72.90 & 68.69 & 70.72 &       & 62.02 & 53.79 & 57.57 &       & 62.23 & 60.21  & 61.19 &       & 69.75 & 69.44 & 69.58 \\
    \midrule

    \textbf{LLM-based} &       &       &       &       &       &       &       &       &       &       &       &       &       &       &  \\
    GPT-3.5-turbo zero-shot  & 44.88     & 55.13     & 49.48 &       & 30.04     & 41.04     & 34.69 &       & 36.02     & 53.40     & 43.02 &       & 39.92     & 57.78     & 47.22 \\
    $\textrm{GPT-3.5-turbo few-shot}$\  & 51.51     & 65.19     & 57.55 &       & 39.79     & 50.09     & 44.35 &       & 43.34     & 63.09     & 51.39 &       & 51.12     & 71.01     & 59.45 \\
    $\textrm{GPT-3.5-turbo CoT}$\  & 48.47     & 59.05     & 53.24 &       & 30.48     & 40.30     & 34.71 &       & 39.51     & 56.70     & 46.57 &       & 44.03     & 63.81     & 52.10 \\
    $\textrm{GPT-3.5-turbo CoT+few-shot}$\  & 49.41     & 59.15     & 53.85 &       & 33.78     & 42.33     & 37.57 &       & 39.02     & 56.08     & 46.02 &       & 46.49     & 66.93     & 54.86 \\
    \textrm{GPT-4o zero-shot}  & 32.99 & 38.13 & 35.37 &       & 17.81 & 22.55 & 19.90 &       & 27.85 & 37.73 & 32.05 &       & 32.17 & 43.00 & 36.80 \\
    $\textrm{GPT-4o few-shot}$\  & 54.11 & 66.20 & 59.55 &       & 38.23 & 48.61 & 42.80 &       & 45.57 & 60.41 & 51.95 &       & 52.90 & 71.01 & 60.63 \\
    \textrm{GPT-4o CoT}  & 41.21 & 53.32 & 46.49 &       & 26.98 & 37.71 & 31.46 &       & 33.07 & 50.93 & 40.10 &       & 39.14 & 58.17 & 46.79 \\
    $\textrm{GPT-4o CoT+few-shot}$\  & 46.81 & 59.86 & 52.54 &       & 29.71 & 40.85 & 34.40 &       & 35.08 & 53.81 & 42.47 &       & 41.53 & 61.09 & 49.45 \\
    \midrule
    
    \textbf{Ours} &       &       &       &       &       &       &       &       &       &       &       &       &       &       &  \\
       MiniConGTS   & 76.1  & \textbf{75.08} & \textbf{75.59} &       & 66.82 & \textbf{60.68} & \textbf{63.61}&       & 66.50 & \textbf{63.86} & \underline{65.15} &       & 75.52 & \textbf{74.14} & \textbf{74.83} \\
    \bottomrule
    \hline
    \end{tabular}%
    }
\caption{Experimental results on $\mathcal{D}_2$ \cite{xu2020position}. 
The best results are highlighted in bold,  while the second best results are underscored. 
The results with $\dagger$ are retrieved from \cite{yu2021aspect}. 
The results with $\natural$ are retrieved from \cite{xu2020position}. 
The results with $\sharp$ are retrieved from \cite{peng2020knowing}. 
The results with $\ddagger$ are retrieved from \cite{mao2021joint}. 
}

\label{tab:D2performance}
\end{table*}

We evaluate ASTE performance using the widely accepted \texttt{(Precision, Recall, F1)} metrics. 
Results of the dataset $\mathcal{D}_2$ are shown in Table \ref{tab:D2performance}, while the results of $\mathcal{D}_1$ are presented in Appendix Table \ref{tab:D1performance}. The best results are highlighted in bold, and the second-best results are underscored. Our proposed method consistently achieves state-of-the-art performance or ranks second in most evaluated cases.

Notably, on dataset $\mathcal{D}_1$, the proposed method achieves a substantial 3.08\% improvement in F1 score on the 14Lap subset. This improvement is particularly significant given that the highest score on this subset is the lowest among all datasets, showcasing our method's effectiveness in handling challenging instances. Moreover, on the 14Res subset, our F1 score exceeds 76.00, which, to the best of our knowledge, is the highest reported performance. For dataset $\mathcal{D}_2$, our method outperforms all state-of-the-art approaches by over 1 percentage point on the 14Res, 14Lap, and 16Res subsets. Only on the 16Res subset does the BDTF method \citep{zhang2022boundary} achieve a slightly better performance. 




\subsubsection{Comparison to GPT}
Our proposed method is based on the Pretrain-Finetuning paradigm, which is increasingly challenged by large language models (LLMs) \cite{kojima2022large, wei2021finetuned}. It is concerned about how the advancing capabilities of LLMs might impact the ASTE task.

When compared to advanced LLMs, the performance and computational efficiency of our method stand out. As shown in Tables \ref{tab:D1performance}, \ref{tab:D2performance}, and \ref{tab:gptExample}, even the state-of-the-art LLM, GPT-4, with its staggering number of parameters, does not achieve satisfactory results for ASTE, even with few-shot learning and Chain-of-Thought (CoT) \cite{wei2022chain} enhancement. Additionally, using LLMs introduces significant computational overhead. 
For more information on experiment setting see Appendix \ref{sec:Details of GPT Experients}. For detailed results see Table \ref{tab:D2performance}, \ref{tab:casestudy}, Appendix \ref{sec:ASTE Performance on D1} and Appendix \ref{sec:Detailed Results of GPT Experiments}.
Note that, fine-tuning LLMs may offer some improvements, but it also risks catastrophic forgetting \cite{shi2024continual} and is left for future work. 

To our knowledge, this is the first formal study to evaluate GPT-4's performance on these ASTE datasets, providing valuable insights for future research.

\subsection{Performance on Other ABSA Tasks}
Our method can also effectively handle other ABSA subtasks, including Aspect Extraction (AE), Opinion Extraction (OE), and Aspect Opinion Pair Extraction (AOPE).
AE aims to extract all the \texttt{(Aspect)} terms, OE aims to extract all the \texttt{Opinion} terms, and AOPE aims to extract all the \texttt{(Aspect, Opinion)} pairs from raw text. 
The results for these tasks are presented in Appendix \ref{sec:Performance on Other ABSA Tasks}, where our method consistently achieves best F1-scores across nearly all tasks.

\begin{table}[htbp]
  \centering
  \scalebox{0.52}{
    \begin{tabular}{lccccccccc}
    \hline
    \toprule
    \multirow{2}[4]{*}{\textbf{Models}} & \multicolumn{4}{c}{$\mathcal{D}_1$}        &       & \multicolumn{4}{c}{$\mathcal{D}_2$} \\
\cmidrule{2-5}\cmidrule{7-10}          & \textbf{14Res} & \textbf{14Lap} & \textbf{15Res} & \textbf{16Res} &       & \textbf{14Res} & \textbf{14Lap} & \textbf{15Res} & \textbf{16Res} \\
    \midrule
    \midrule
    \textbf{MiniConGTS}  & 76.00    & 64.07  & 65.43 & 71.80 &       & 75.59 & 63.61 & 65.15 & 74.83 \\
    \midrule
        w/o. RoBERTa & 74.12 & 63.18 & 62.95 & 69.41 &       & 72.66 & 62.15 & 63.25 & 70.71 \\
    \ \ \ \ $\Delta$F\_1  & -1.88 & -0.89 & -2.48  & -2.39  &       & -2.93 & -1.46 & -1.90    & -4.12 \\
    \midrule
        \ \ \ \ w/o. contr & 72.61 & 61.94 & 58.14 & 68.16 &       & 71.72 & 61.49 & 58.11 & 68.03 \\
    \ \ \ \ $\Delta$F\_1  & -3.39 & -2.13 & -7.29 & -3.64 &       & -3.87 & -2.12 & -7.04 & -6.80 \\
    \midrule
        \ \ \ \ w/o. tag & 67.78 & 54.98 & 60.75 & 62.62 &       & 65.83 & 54.98 & 58.73 & 67.63 \\
    \ \ \ \ $\Delta$F\_1  & -8.22 & -9.09 & -4.68 & -9.18 &       & -9.76 & -8.63 & -6.42 & -7.20 \\
    \bottomrule
    \hline
    \end{tabular}%
    }
    \caption{Ablation study on 
    F1, 
    where ``w/o. RoBERTa'' denotes ``Replace RoBERTa with bert-base-uncased'', ``w/o. contr'' denotes without the contrastive learning mechanism, and ``w/o. tag'' denotes ``replace our tagging scheme with a baseline''.}
    \label{ablation}%
\end{table}%

\begin{table}[htbp]
  \centering

  \scalebox{0.62}{
    \begin{tabular}{ccc}
    \toprule
    Method & Num Tags& Features Enhancing\\
    \midrule
    \midrule
    GTS \cite{wu2020grid} & 6 & None\\
    Double-encoder \cite{jing2021seeking} & 9 & None\\
    EMC-GCN \cite{chen2022enhanced} & 10 & 4 Groups \\
    BDTF \cite{zhang2022boundary} & $2\times2\times3$ & None\\
    STAGE \cite{liang2023stage} & $2\times2\times4$ & None\\
    DGCNAP \cite{li2023dual} & 6 & POS-tagging\\
    \midrule
    Ours  & $5$ & None    \\
    \bottomrule
    \end{tabular}%
      }
        \caption{Tagging Scheme Comparison. }
  \label{tab:taggingCompare}%
\end{table}%

\section{Analysis}
\subsection{Ablation Study}
In this section, we conduct a series of ablation experiments to demonstrate the superiority of our method and eliminate potential confounding factors. Experiments were conducted on the $\mathcal{D}_1$ and $\mathcal{D}_2$ datasets, using F1 scores as the comparison metric.

\noindent\textbf{Encoder}. We replaced the RoBERTa encoder with BERT, resulting in a slight decrease in F1 scores on both datasets, although our method still outperformed most other approaches.

\noindent\textbf{Contrastive Learning}. We deactivated the contrastive mechanism in our method (denoted as ``w/o. contr'') by setting the coefficient of the contrastive loss to 0. The results in Table \ref{ablation} illustrate a significant F1-score decrease of $2.12\sim 7.29$\% in both datasets. 


\noindent\textbf{Tagging Scheme}. We substituted our proposed scheme with the conventional GTS tagging scheme \cite{wu2020grid}, resulting in a substantial performance decline (Table \ref{ablation}) of \textcolor{black}{$4.68\sim9.18$\%}. 
This indicates that the contrastive learning methods, within our framework, is of strong reliance on an appropriate tagging scheme. This reinforces the effectiveness of our compact yet impactful tagging scheme.

\subsection{Effect of Contrastive Learning}
\begin{figure*}[ht]
    \centering
    \includegraphics[width=\linewidth]{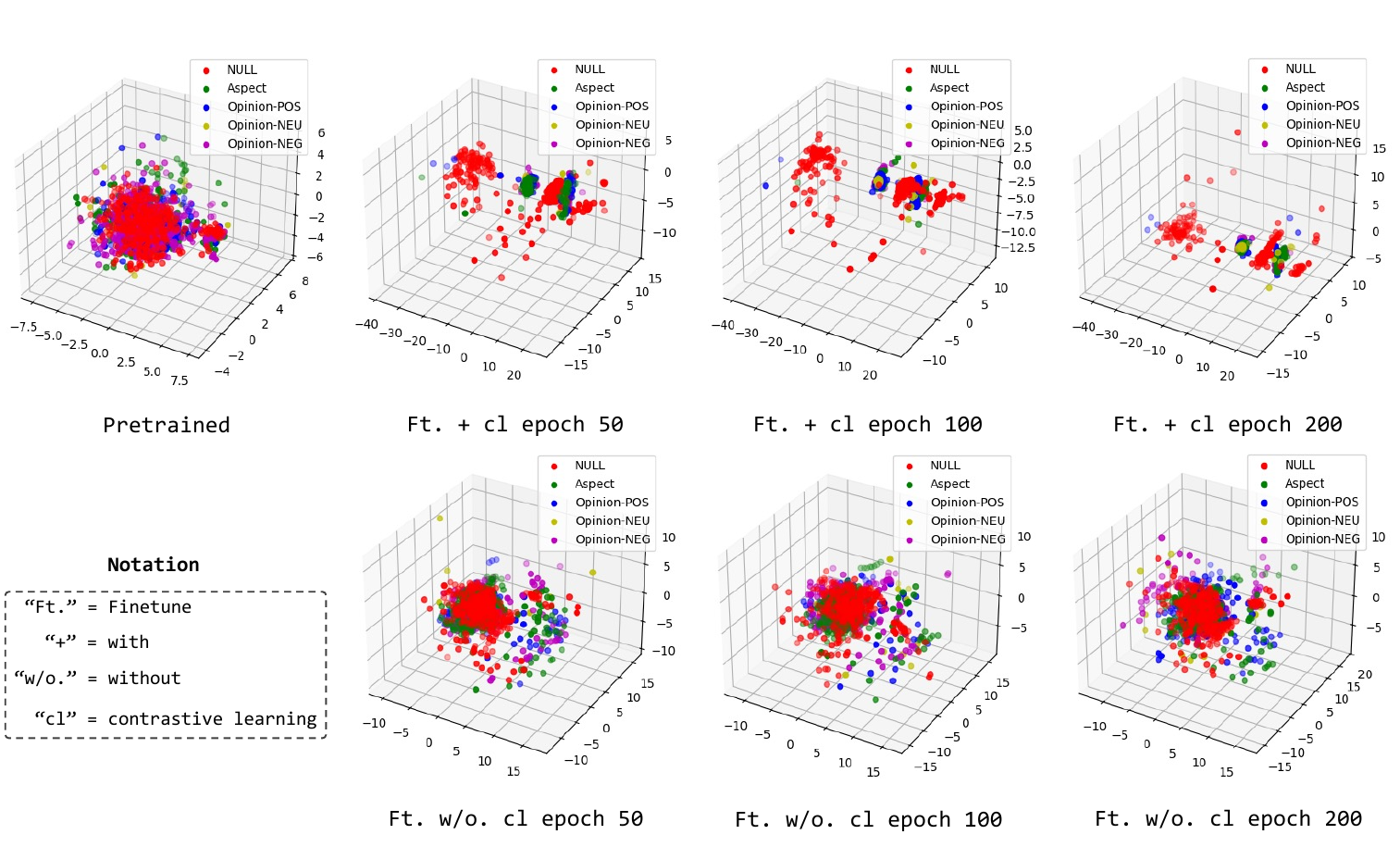}
    \caption{A plot of the hidden word representation based on the $\mathcal{D}_1$ 14Res dataset, where the dimension is reduced to 3. ``Pretrained'' refers to the representation output by official released model. We finetune the pretrained model with and without contrastive learning strategy respectively. }
    \label{fig:fig_contr_plot.pdf}
\end{figure*}

In Figure \ref{fig:fig_contr_plot.pdf}, an example is shown illustrating how contrastive learning improves representation. The right upper-row subplots show the representation outputs with contrastive learning, while the lower row subplots display that without contrastive learning. Principal Component Analysis (PCA) \citep{mackiewicz1993principal} is used to reduce the vector dimensions to three for visualization purposes. The distributions indicate that contrastive learning significantly enhances the representations, with similar classes of hidden word representations becoming more tightly clustered and dissimilar classes more distinct.




\subsection{Efficiency Comparison}

\begin{table*}[ht]
  \centering
  
  \scalebox{0.8}{
    \begin{tabular}{ccccccc}
    \toprule
    Model & Memory & Num Params & Epoch Time$^\sharp$ & Inf Time & F1(\%) & Device \\
    \midrule
    Span-ASTE \cite{xu2021learning} & 3.173 GB$^\flat$ & - & 108s & - & 71.62 & Tesla v100 32GB\\
    BDTF \cite{zhang2022boundary} & 8.103 GB$^\flat$ & $>$0.18B$^\flat$ & 135s & - & 74.73 & Tesla v100 32GB \\
    GPT 3.5-Turbo \cite{openai_api}& $>$80 GB$^\natural$ & 175B$^\dagger$ & - & 0.83s & 49.48 & OpenAI API \\
    GPT 4 \cite{openai_api}& $>$80 GB$^\natural$ & 1760B$^\ddagger$ & - & 1.56s & 35.37 & OpenAI API \\
    \midrule
    Ours  & 7.11GB & 0.12B & 10s & 0.01s &  76.00    & 2080 Ti 11GB\\
    \bottomrule
    \end{tabular}%
      }
\caption{An efficiency comparison, where $^\dagger$ is evaluated by \cite{gao2021sizes} and later confirmed by OpenAI \cite{GPT3Wiki64:online}, $^\ddagger$ is estimated by \cite{schreiner2023gpt4}, $^\flat$ is cited from \cite{zhang2022boundary}, and $^\natural$ is reported by \cite{GPT3Wiki64:online}. $^\sharp$ Epoch Time  refers to the training time per epoch on the training set. }
  \label{tab:efficiencyComparison}%

\end{table*}%



We compared the computational efficiency of MiniConGTS with other approaches, including baseline ASTE methods and LLMs, on an ASTE task. Evaluation metrics such as memory usage, number of parameters, epoch/inference time, and F1 scores are recorded in Table \ref{tab:efficiencyComparison}. Our approach not only requires less memory usage for higher performance compared to traditional ASTE methods but also offers much faster runtime even using a relatively lower-cost GPU. 


Table \ref{tab:taggingCompare} provides another comparative analysis of tagging schemes. Our method has a compact design with the fewest classes of labels. What's more, compared with other SOTA baselines, our method does not rely on any additional linguistic information enhancement.


\subsection{Case Study}
A case analysis is presented in Table \ref{tab:casestudy}, where the proposed method demonstrates solid performance. Despite minor faults in missing the full terms, it exhibits a profound understanding of the case. 

It is quite interesting to investigate the GPT's error cases. 
The findings reveal that the performance of the GPT model is mixed - while it is able to identify more aspect-opinion (A-O) pairs than the ground truth annotations in some cases, this comes at the cost of a decreased precision. This suggests that the GPT model may be ``over-interpreting'' the input, making inferences that go beyond what is strictly supported by the text.
Furthermore, the GPT model appears to be overly sensitive to the presence of adverbs (such as ``very'', ``a bit'', etc.) in the input. This sensitivity manifests in the model frequently adding or removing adverbs when extracting the Opinion components, which further contributes to a decrease in the overall accuracy of the ABSA task.

These findings highlight the importance of developing ABSA models that can strike the right balance between extracting all relevant aspect-opinion pairs, while still maintaining a high degree of precision. The effective use of encoding appears to be a promising direction for achieving this balance and advancing the state-of-the-art in Aspect-Based Sentiment Analysis.




\section{Conclusion}
In this work, we have introduced an elegant and efficient framework for ASTE, achieving SOTA performance. Our approach is built upon two effective components: a new tagging scheme and a novel token-level contrastive learning implementation. The ablation study demonstrates the synergy between these components, reducing the need for complex model designs and external information enhancements. 

\section{Limitations}
Our method is based on a 2D-matrix tagging scheme, where the time complexity for decoding, given the input corpus length \(N\), is \(O(N^2)\). This may be unacceptable when \(N\) is too large. 
Additionally, although we have demonstrated our method on commonly used classic English datasets, it should be tested on more natural corpora and for its cross-language capability.

\section{Ethics \& Potential Risks Statement}
In our experiments, we used widely accepted datasets focused on e-commerce reviews, which have a lower risk of offensive content. We scrutinized the data for biases against gender, race, and marginalized groups. Despite these precautions, our model might still generate potentially offensive sentiment assessments if used inappropriately, such as evaluating ethical or moral statements. We reserve the right to limit or modify the use of our technology to prevent misuse.


\section{Acknowledgement}
This work was partially supported by the National Key R\&D Program of China under Grant No. 2022ZD0160101. The authors would also like to thank Shanghai Artificial Intelligence Laboratory for their support and resources, as well as colleagues from Shanghai Jiao Tong University for valuable discussions. This work was conducted during the internships of Qiao Sun, Liujia Yang, and Minghao Ma at the Shanghai Artificial Intelligence Laboratory.

\bibliography{emnlp2023}

\begin{thebibliography}{47}
\expandafter\ifx\csname natexlab\endcsname\relax\def\natexlab#1{#1}\fi

\bibitem[{Chen et~al.(2022)Chen, Zhai, Feng, Li, and Wang}]{chen2022enhanced}
Hao Chen, Zepeng Zhai, Fangxiang Feng, Ruifan Li, and Xiaojie Wang. 2022.
\newblock Enhanced multi-channel graph convolutional network for aspect sentiment triplet extraction.
\newblock In \emph{Proceedings of the 60th Annual Meeting of the Association for Computational Linguistics (Volume 1: Long Papers)}, pages 2974--2985.

\bibitem[{Chen et~al.(2021{\natexlab{a}})Chen, Wang, Liu, and Wang}]{chen2021bidirectional}
Shaowei Chen, Yu~Wang, Jie Liu, and Yuelin Wang. 2021{\natexlab{a}}.
\newblock Bidirectional machine reading comprehension for aspect sentiment triplet extraction.
\newblock In \emph{Proceedings of the AAAI conference on artificial intelligence}, volume~35, pages 12666--12674.

\bibitem[{Chen et~al.(2021{\natexlab{b}})Chen, Huang, Liu, Shi, and Jin}]{chen2021semantic}
Zhexue Chen, Hong Huang, Bang Liu, Xuanhua Shi, and Hai Jin. 2021{\natexlab{b}}.
\newblock Semantic and syntactic enhanced aspect sentiment triplet extraction.
\newblock \emph{arXiv preprint arXiv:2106.03315}.

\bibitem[{Dai and Song(2019)}]{dai2019neural}
Hongliang Dai and Yangqiu Song. 2019.
\newblock Neural aspect and opinion term extraction with mined rules as weak supervision.
\newblock In \emph{Proceedings of the 57th Annual Meeting of the Association for Computational Linguistics}, pages 5268--5277.

\bibitem[{Fei et~al.(2022)Fei, Li, Li, Wu, Li, and Ji}]{fei2022inheriting}
Hao Fei, Fei Li, Chenliang Li, Shengqiong Wu, Jingye Li, and Donghong Ji. 2022.
\newblock Inheriting the wisdom of predecessors: A multiplex cascade framework for unified aspect-based sentiment analysis.
\newblock In \emph{Proceedings of the Thirty-First International Joint Conference on Artificial Intelligence, IJCAI}, pages 4096--4103.

\bibitem[{Gao(2021)}]{gao2021sizes}
Leo Gao. 2021.
\newblock On the sizes of openai api models.
\newblock \url{https://blog.eleuther.ai/gpt3-model-sizes/}.
\newblock Retrieved November 23, 2023.

\bibitem[{Gao et~al.(2021)Gao, Yao, and Chen}]{gao2021simcse}
Tianyu Gao, Xingcheng Yao, and Danqi Chen. 2021.
\newblock Simcse: Simple contrastive learning of sentence embeddings.
\newblock In \emph{2021 Conference on Empirical Methods in Natural Language Processing, EMNLP 2021}, pages 6894--6910. Association for Computational Linguistics (ACL).

\bibitem[{Giorgi et~al.(2021)Giorgi, Nitski, Wang, and Bader}]{giorgi2021declutr}
John Giorgi, Osvald Nitski, Bo~Wang, and Gary Bader. 2021.
\newblock Declutr: Deep contrastive learning for unsupervised textual representations.
\newblock In \emph{Proceedings of the 59th Annual Meeting of the Association for Computational Linguistics and the 11th International Joint Conference on Natural Language Processing (Volume 1: Long Papers)}, pages 879--895.

\bibitem[{Hinton and Salakhutdinov(2006)}]{Hinton2006}
Geoffrey~E. Hinton and Ruslan~R. Salakhutdinov. 2006.
\newblock Reducing the dimensionality of data with neural networks.
\newblock \emph{Science}, 313(5786):504--507.

\bibitem[{Iswari et~al.(2024)Iswari, Afriliana, Dharma, and Yuniari}]{iswari2024enhancing}
Ni~Made~Satvika Iswari, Nunik Afriliana, Eddy~Muntina Dharma, and Ni~Putu~Widya Yuniari. 2024.
\newblock Enhancing aspect-based sentiment analysis in visitor review using semantic similarity.
\newblock \emph{Journal of Applied Data Sciences}, 5(2):724--735.

\bibitem[{Jiang et~al.(2023)Jiang, Liang, Liu, Dong, and Li}]{jiang2023semantically}
Baoxing Jiang, Shehui Liang, Peiyu Liu, Kaifang Dong, and Hongye Li. 2023.
\newblock A semantically enhanced dual encoder for aspect sentiment triplet extraction.
\newblock \emph{Neurocomputing}, 562:126917.

\bibitem[{Jing et~al.(2021)Jing, Li, Zhao, and Jiang}]{jing2021seeking}
Hongjiang Jing, Zuchao Li, Hai Zhao, and Shu Jiang. 2021.
\newblock Seeking common but distinguishing difference, a joint aspect-based sentiment analysis model.
\newblock In \emph{Proceedings of the 2021 Conference on Empirical Methods in Natural Language Processing}, pages 3910--3922.

\bibitem[{Kojima et~al.(2022)Kojima, Gu, Reid, Matsuo, and Iwasawa}]{kojima2022large}
Takeshi Kojima, Shixiang~Shane Gu, Machel Reid, Yutaka Matsuo, and Yusuke Iwasawa. 2022.
\newblock Large language models are zero-shot reasoners.
\newblock \emph{Advances in neural information processing systems}, 35:22199--22213.

\bibitem[{Li et~al.(2023)Li, He, and Zhang}]{li2023dual}
Yanbo Li, Qing He, and Damin Zhang. 2023.
\newblock Dual graph convolutional networks integrating affective knowledge and position information for aspect sentiment triplet extraction.
\newblock \emph{Frontiers in Neurorobotics}, 17.

\bibitem[{Liang et~al.(2023)Liang, Wei, Mao, Fu, Fang, and Chen}]{liang2023stage}
Shuo Liang, Wei Wei, Xian-Ling Mao, Yuanyuan Fu, Rui Fang, and Dangyang Chen. 2023.
\newblock Stage: span tagging and greedy inference scheme for aspect sentiment triplet extraction.
\newblock In \emph{Proceedings of the AAAI Conference on Artificial Intelligence}, volume~37, pages 13174--13182.

\bibitem[{Lin et~al.(2017)Lin, Goyal, Girshick, He, and Doll{\'a}r}]{lin2017focal}
Tsung-Yi Lin, Priya Goyal, Ross Girshick, Kaiming He, and Piotr Doll{\'a}r. 2017.
\newblock Focal loss for dense object detection.
\newblock In \emph{Proceedings of the IEEE international conference on computer vision}, pages 2980--2988.

\bibitem[{Ma{\'c}kiewicz and Ratajczak(1993)}]{mackiewicz1993principal}
Andrzej Ma{\'c}kiewicz and Waldemar Ratajczak. 1993.
\newblock Principal components analysis (pca).
\newblock \emph{Computers \& Geosciences}, 19(3):303--342.

\bibitem[{Mao et~al.(2021)Mao, Shen, Yu, and Cai}]{mao2021joint}
Yue Mao, Yi~Shen, Chao Yu, and Longjun Cai. 2021.
\newblock A joint training dual-mrc framework for aspect based sentiment analysis.
\newblock In \emph{Proceedings of the AAAI conference on artificial intelligence}, volume~35, pages 13543--13551.

\bibitem[{{OpenAI}(2024)}]{openai_api}
{OpenAI}. 2024.
\newblock {OpenAI API}.
\newblock \url{https://platform.openai.com/docs}.
\newblock Accessed: 2024-04-14.

\bibitem[{Peng et~al.(2020)Peng, Xu, Bing, Huang, Lu, and Si}]{peng2020knowing}
Haiyun Peng, Lu~Xu, Lidong Bing, Fei Huang, Wei Lu, and Luo Si. 2020.
\newblock Knowing what, how and why: A near complete solution for aspect-based sentiment analysis.
\newblock In \emph{Proceedings of the AAAI conference on artificial intelligence}, volume~34, pages 8600--8607.

\bibitem[{Pontiki et~al.(2014)Pontiki, Hadjipavlou-Litina, Litinas, and Geromichalos}]{pontiki2014novel}
Eleni Pontiki, Dimitra Hadjipavlou-Litina, Konstantinos Litinas, and George Geromichalos. 2014.
\newblock Novel cinnamic acid derivatives as antioxidant and anticancer agents: Design, synthesis and modeling studies.
\newblock \emph{Molecules}, 19(7):9655--9674.

\bibitem[{Pontiki et~al.(2015)Pontiki, Galanis, Papageorgiou, Manandhar, and Androutsopoulos}]{pontiki2015semeval}
Maria Pontiki, Dimitrios Galanis, Harris Papageorgiou, Suresh Manandhar, and Ion Androutsopoulos. 2015.
\newblock Semeval-2015 task 12: Aspect based sentiment analysis.
\newblock In \emph{Proceedings of the 9th international workshop on semantic evaluation (SemEval 2015)}, pages 486--495.

\bibitem[{Pontiki et~al.(2016)Pontiki, Galanis, Papageorgiou, Androutsopoulos, Manandhar, AL-Smadi, Al-Ayyoub, Zhao, Qin, De~Clercq et~al.}]{pontiki2016semeval}
Maria Pontiki, Dimitris Galanis, Haris Papageorgiou, Ion Androutsopoulos, Suresh Manandhar, Mohammed AL-Smadi, Mahmoud Al-Ayyoub, Yanyan Zhao, Bing Qin, Orph{\'e}e De~Clercq, et~al. 2016.
\newblock Semeval-2016 task 5: Aspect based sentiment analysis.
\newblock In \emph{ProWorkshop on Semantic Evaluation (SemEval-2016)}, pages 19--30. Association for Computational Linguistics.

\bibitem[{Schreiner(2023)}]{schreiner2023gpt4}
Maximilian Schreiner. 2023.
\newblock Gpt-4 architecture, datasets, costs and more leaked.
\newblock THE DECODER.
\newblock Archived from the original on July 12, 2023. Retrieved July 12, 2023.

\bibitem[{Shi et~al.(2024)Shi, Xu, Wang, Qin, Wang, Wang, and Wang}]{shi2024continual}
Haizhou Shi, Zihao Xu, Hengyi Wang, Weiyi Qin, Wenyuan Wang, Yibin Wang, and Hao Wang. 2024.
\newblock Continual learning of large language models: A comprehensive survey.
\newblock \emph{arXiv preprint arXiv:2404.16789}.

\bibitem[{van~den Oord et~al.(2019)van~den Oord, Li, and Vinyals}]{oord2019representation}
Aaron van~den Oord, Yazhe Li, and Oriol Vinyals. 2019.
\newblock \href {http://arxiv.org/abs/1807.03748} {Representation learning with contrastive predictive coding}.

\bibitem[{Wang et~al.(2022)Wang, Ding, Zhong, Li, and Tao}]{wang2022contrastive}
Bing Wang, Liang Ding, Qihuang Zhong, Ximing Li, and Dacheng Tao. 2022.
\newblock A contrastive cross-channel data augmentation framework for aspect-based sentiment analysis.
\newblock In \emph{Proceedings of the 29th International Conference on Computational Linguistics}, pages 6691--6704.

\bibitem[{Wang et~al.(2017)Wang, Pan, Dahlmeier, and Xiao}]{wang2017coupled}
Wenya Wang, Sinno~Jialin Pan, Daniel Dahlmeier, and Xiaokui Xiao. 2017.
\newblock Coupled multi-layer attentions for co-extraction of aspect and opinion terms.
\newblock In \emph{Proceedings of the Thirty-First AAAI Conference on Artificial Intelligence}, pages 3316--3322.

\bibitem[{Wei et~al.(2021)Wei, Bosma, Zhao, Guu, Yu, Lester, Du, Dai, and Le}]{wei2021finetuned}
Jason Wei, Maarten Bosma, Vincent~Y Zhao, Kelvin Guu, Adams~Wei Yu, Brian Lester, Nan Du, Andrew~M Dai, and Quoc~V Le. 2021.
\newblock Finetuned language models are zero-shot learners.
\newblock \emph{arXiv preprint arXiv:2109.01652}.

\bibitem[{Wei et~al.(2022)Wei, Wang, Schuurmans, Bosma, Xia, Chi, Le, Zhou et~al.}]{wei2022chain}
Jason Wei, Xuezhi Wang, Dale Schuurmans, Maarten Bosma, Fei Xia, Ed~Chi, Quoc~V Le, Denny Zhou, et~al. 2022.
\newblock Chain-of-thought prompting elicits reasoning in large language models.
\newblock \emph{Advances in neural information processing systems}, 35:24824--24837.

\bibitem[{Wikipedia(2024)}]{GPT3Wiki64:online}
Wikipedia. 2024.
\newblock Gpt-3 - wikipedia.
\newblock \url{https://en.wikipedia.org/wiki/GPT-3}.
\newblock (Accessed on 04/14/2024).

\bibitem[{Wolf et~al.(2020)Wolf, Debut, Sanh, Chaumond, Delangue, Moi, Cistac, Rault, Louf, Funtowicz et~al.}]{wolf2020transformers}
Thomas Wolf, Lysandre Debut, Victor Sanh, Julien Chaumond, Clement Delangue, Anthony Moi, Pierric Cistac, Tim Rault, R{\'e}mi Louf, Morgan Funtowicz, et~al. 2020.
\newblock Transformers: State-of-the-art natural language processing.
\newblock In \emph{Proceedings of the 2020 conference on empirical methods in natural language processing: system demonstrations}, pages 38--45.

\bibitem[{Wu et~al.(2020{\natexlab{a}})Wu, Ying, Zhao, Fan, Dai, and Xia}]{wu2020grid}
Zhen Wu, Chengcan Ying, Fei Zhao, Zhifang Fan, Xinyu Dai, and Rui Xia. 2020{\natexlab{a}}.
\newblock Grid tagging scheme for aspect-oriented fine-grained opinion extraction.
\newblock In \emph{Findings of the Association for Computational Linguistics: EMNLP 2020}, pages 2576--2585.

\bibitem[{Wu et~al.(2020{\natexlab{b}})Wu, Wang, Gu, Khabsa, Sun, and Ma}]{wu2020clear}
Zhuofeng Wu, Sinong Wang, Jiatao Gu, Madian Khabsa, Fei Sun, and Hao Ma. 2020{\natexlab{b}}.
\newblock Clear: Contrastive learning for sentence representation.
\newblock \emph{arXiv preprint arXiv:2012.15466}.

\bibitem[{Xu et~al.(2021)Xu, Chia, and Bing}]{xu2021learning}
Lu~Xu, Yew~Ken Chia, and Lidong Bing. 2021.
\newblock Learning span-level interactions for aspect sentiment triplet extraction.
\newblock \emph{arXiv preprint arXiv:2107.12214}.

\bibitem[{Xu et~al.(2020)Xu, Li, Lu, and Bing}]{xu2020position}
Lu~Xu, Hao Li, Wei Lu, and Lidong Bing. 2020.
\newblock Position-aware tagging for aspect sentiment triplet extraction.
\newblock \emph{arXiv preprint arXiv:2010.02609}.

\bibitem[{Yan et~al.(2021)Yan, Dai, Ji, Qiu, and Zhang}]{yan2021unified}
Hang Yan, Junqi Dai, Tuo Ji, Xipeng Qiu, and Zheng Zhang. 2021.
\newblock A unified generative framework for aspect-based sentiment analysis.
\newblock In \emph{Proceedings of the 59th Annual Meeting of the Association for Computational Linguistics and the 11th International Joint Conference on Natural Language Processing (Volume 1: Long Papers)}, pages 2416--2429.

\bibitem[{Yang et~al.(2023)Yang, Zhang, Hu, and Zhou}]{yang2023pairing}
Fan Yang, Mian Zhang, Gongzhen Hu, and Xiabing Zhou. 2023.
\newblock A pairing enhancement approach for aspect sentiment triplet extraction.
\newblock \emph{arXiv preprint arXiv:2306.10042}.

\bibitem[{Ye et~al.(2021)Ye, Zhang, Deng, Chen, Tan, Huang, and Chen}]{ye2021contrastive}
Hongbin Ye, Ningyu Zhang, Shumin Deng, Mosha Chen, Chuanqi Tan, Fei Huang, and Huajun Chen. 2021.
\newblock Contrastive triple extraction with generative transformer.
\newblock In \emph{Proceedings of the AAAI conference on artificial intelligence}, volume~35, pages 14257--14265.

\bibitem[{Yu~Bai~Jian et~al.(2021)Yu~Bai~Jian, Nayak, Majumder, and Poria}]{yu2021aspect}
Samson Yu~Bai~Jian, Tapas Nayak, Navonil Majumder, and Soujanya Poria. 2021.
\newblock Aspect sentiment triplet extraction using reinforcement learning.
\newblock In \emph{Proceedings of the 30th ACM International Conference on Information \& Knowledge Management}, pages 3603--3607.

\bibitem[{Zhai et~al.(2022)Zhai, Chen, Feng, Li, and Wang}]{zhai2022mrc}
Zepeng Zhai, Hao Chen, Fangxiang Feng, Ruifan Li, and Xiaojie Wang. 2022.
\newblock Com-mrc: A context-masked machine reading comprehension framework for aspect sentiment triplet extraction.
\newblock In \emph{Proceedings of the 2022 Conference on Empirical Methods in Natural Language Processing}, pages 3230--3241.

\bibitem[{Zhang et~al.(2020)Zhang, Li, Song, and Wang}]{zhang2020multi}
Chen Zhang, Qiuchi Li, Dawei Song, and Benyou Wang. 2020.
\newblock A multi-task learning framework for opinion triplet extraction.
\newblock In \emph{Findings of the Association for Computational Linguistics: EMNLP 2020}, pages 819--828.

\bibitem[{Zhang et~al.(2021)Zhang, Nan, Wei, Li, Zhu, Mckeown, Nallapati, Arnold, and Xiang}]{zhang2021supporting}
Dejiao Zhang, Feng Nan, Xiaokai Wei, Shang-Wen Li, Henghui Zhu, Kathleen Mckeown, Ramesh Nallapati, Andrew~O Arnold, and Bing Xiang. 2021.
\newblock Supporting clustering with contrastive learning.
\newblock In \emph{Proceedings of the 2021 Conference of the North American Chapter of the Association for Computational Linguistics: Human Language Technologies}, pages 5419--5430.

\bibitem[{Zhang et~al.(2022{\natexlab{a}})Zhang, Li, Deng, Bing, and Lam}]{zhang2022survey}
Wenxuan Zhang, Xin Li, Yang Deng, Lidong Bing, and Wai Lam. 2022{\natexlab{a}}.
\newblock A survey on aspect-based sentiment analysis: Tasks, methods, and challenges.
\newblock \emph{IEEE Transactions on Knowledge and Data Engineering}.

\bibitem[{Zhang et~al.(2022{\natexlab{b}})Zhang, Yang, Li, Liang, Chen, Dang, Yang, and Xu}]{zhang2022boundary}
Yice Zhang, Yifan Yang, Yihui Li, Bin Liang, Shiwei Chen, Yixue Dang, Min Yang, and Ruifeng Xu. 2022{\natexlab{b}}.
\newblock Boundary-driven table-filling for aspect sentiment triplet extraction.
\newblock In \emph{Proceedings of the 2022 Conference on Empirical Methods in Natural Language Processing}, pages 6485--6498.

\bibitem[{Zheng et~al.(2017)Zheng, Wang, Bao, Hao, Zhou, and Xu}]{zheng2017joint}
Suncong Zheng, Feng Wang, Hongyun Bao, Yuexing Hao, Peng Zhou, and Bo~Xu. 2017.
\newblock Joint extraction of entities and relations based on a novel tagging scheme.
\newblock \emph{arXiv preprint arXiv:1706.05075}.

\bibitem[{Zou et~al.(2024)Zou, Zhang, Wu, and Tian}]{zou2024multi}
Wang Zou, Wubo Zhang, Wenhuan Wu, and Zhuoyan Tian. 2024.
\newblock A multi-task shared cascade learning for aspect sentiment triplet extraction using bert-mrc.
\newblock \emph{Cognitive Computation}, pages 1--18.

\end{thebibliography}
\bibliographystyle{acl_natbib}

\appendix

\section{Appendix}
\label{sec:appendix}

\subsection{Pseudo-code for the training process.}
See Algorithm \ref{alg:training}.
\label{sec:pseudo}
\begin{algorithm}[!ht]
\small

\caption{.}

\textbf{Modules}:









\textbf{Input}: 

\ \ Raw sentences: $\mathcal{S}_{|\mathcal{S}|}$; 

\ \ Ground truth triplets: $\mathcal{T}_{|\mathcal{T}|}^{gt}$ , where

\ \ \ $\mathcal{T}_k=(A_k, O_k, S_k)$, $k \in \{1,2,...,|\mathcal{T}|\}$;

\ \  classes of contrasted labels: $\mathcal{C}$.

\textbf{Output}: 



\ \ Predicted Triplets: $\mathcal{T}_{|\mathcal{T}|}^{pred}$;

\ \ Metric: $\mathit{Precision}, \mathit{Recall}, \mathit{F1}$.

\textbf{Algorithm}:\\
Repeat for $N$ epochs:

\begin{algorithmic}[1] 
\STATE Hidden word representation: \\ $\mathcal{H}_{|\mathcal{H}|} = \texttt{PLMsEncoder}(\mathcal{S}_{|\mathcal{S}|})$;
\STATE Tensor Operations: 

 $\mathcal{H}_{|\mathcal{H}|\times |\mathcal{H}|}=\texttt{expand}(\mathcal{H}_{|\mathcal{H}|})$, 

 $\mathcal{H}_{|\mathcal{H}|\times |\mathcal{H}|}^T=\mathcal{H}_{|\mathcal{H}|\times |\mathcal{H}|}.\texttt{transpose}()$;
\STATE Similarity matrix: 

$\mathbf{Sim}_{|\mathcal{H}|\times |\mathcal{H}|}=$

$- (\mathcal{H}_{|\mathcal{H}|\times |\mathcal{H}|}-\mathcal{H}_{|\mathcal{H}|\times |\mathcal{H}|}^T) \circ (\mathcal{H}_{|\mathcal{H}|\times |\mathcal{H}|}-\mathcal{H}_{|\mathcal{H}|\times |\mathcal{H}|}^T)$

where 
\[
\mathbf{Sim}_{i, j} = -\left\lVert \mathcal{H}_i - \mathcal{H}_j \right\rVert^2
\]
and $\circ$ denotes the Hadamard product.
\STATE Contrastive Mask matrix: $\mathbf{M}_{|\mathcal{H}|\times |\mathcal{H}|}$, where \  $\mathbf{M}_{i, j} = 1$\ if 
$\mathcal{H}_i, mathcal{H}_j\in \mathcal{C}_p, p\in{1, 2 , 3}$ 
else $-1$;

\STATE Contrastive loss: 

$\mathcal{L}_{contrastive}=\newline \sum_{i=1}^{|\mathcal{H}|}\sum_{j=1}^{|\mathcal{H}|}\big(\mathbf{Sim_{|\mathcal{H}|\times|\mathcal{H}|}} \circ \mathbf{M_{|\mathcal{H}|\times|\mathcal{H}|}}\big)_{i, j};$

\STATE Predicted tagging matrix: \\
$\mathbf{Tag}_{|\mathcal{H}|\times |\mathcal{H}|}^{pred}=\texttt{ClsHead}(\mathcal{H}_{|\mathcal{H}|\times |\mathcal{H}|}, \mathcal{H}_{|\mathcal{H}|\times |\mathcal{H}|}^T)$;
\STATE Focal loss: \\
$\mathcal{L}_{focal} = \texttt{FocalLoss}(\mathbf{Tag}_{|\mathcal{H}|\times |\mathcal{H}|}^{pred}, \mathbf{Tag}_{|\mathcal{H}|\times |\mathcal{H}|}^{gt})$;
\STATE Weighted Loss: $\mathcal{L}=\mathcal{L}_{focal} + \alpha \mathcal{L}_{contrastive}$.
\STATE Backward propagation.

\end{algorithmic}

Predicted triplets: 

\ \ $\mathcal{T}_{|\mathcal{T}|}^{pred}=\texttt{TaggingDecoder}(\mathbf{Tag}_{|\mathcal{H}|\times |\mathcal{H}|}^{pred})$

Metric:

\ \ $\mathit{Precision}, \mathit{Recall}, \mathit{F1} = \texttt{Metric}( \mathcal{T}_{|\mathcal{T}|}^{pred}, \mathcal{T}_{|\mathcal{T}|}^{gt})$

\label{alg:training}
\end{algorithm}

\subsection{Descriptive Statistics of The Datasets}
\label{sec:Descriptive Statistics of The Datasets}
See Table \ref{tab:data}.
\begin{table}
  \centering
  \scalebox{0.64}{
    \begin{tabular}{cccccccccc}
    \hline
    \toprule
    \multicolumn{3}{c}{\textbf{Datasets}} & \textbf{\#S} & \textbf{\#A} & \textbf{\#O} & \textbf{\#S1} & \textbf{\#S2 } & \textbf{\#S3} & \textbf{\#T} \\
    \hline
    \midrule
    
    \multirow{6}[4]{*}{\textbf{14Res}} & \multirow{3}[2]{*}{$\mathcal{D}_1$} & \textbf{Train} & 1259  & 1008  & 849   & 1456  & 164   & 446   & 2066 \\
          &       & \textbf{Dev} & 315   & 358   & 321   & 352   & 44    & 93    & 489 \\
          &       & \textbf{Test} & 493   & 591   & 433   & 651   & 59    & 141   & 851 \\
\cmidrule{2-10}          & \multirow{3}[2]{*}{$\mathcal{D}_2$} & \textbf{Train} & 1266  & 986   & 844   & 1692  & 166   & 480   & 2338 \\
          &       & \textbf{Dev} & 310   & 396   & 307   & 404   & 54    & 119   & 577 \\
          &       & \textbf{Test} & 492   & 579   & 437   & 773   & 66    & 155   & 994 \\
    \midrule
    \multirow{6}[4]{*}{\textbf{14Lap}} & \multirow{3}[2]{*}{$\mathcal{D}_1$} & \textbf{Train} & 899   & 731   & 693   & 691   & 107   & 466   & 1264 \\
          &       & \textbf{Dev} & 225   & 303   & 237   & 173   & 42    & 118   & 333 \\
          &       & \textbf{Test} & 332   & 411   & 330   & 305   & 62    & 101   & 468 \\
\cmidrule{2-10}          & \multirow{3}[2]{*}{$\mathcal{D}_2$} & \textbf{Train} & 906   & 733   & 695   & 817   & 126   & 517   & 1460 \\
          &       & \textbf{Dev} & 219   & 268   & 237   & 169   & 36    & 141   & 346 \\
          &       & \textbf{Test} & 328   & 400   & 329   & 364   & 63    & 116   & 543 \\
    \midrule
    \multirow{6}[4]{*}{\textbf{15Res}} & \multirow{3}[2]{*}{$\mathcal{D}_1$} & \textbf{Train} & 603   & 585   & 485   & 668   & 24    & 179   & 871 \\
          &       & \textbf{Dev} & 151   & 182   & 161   & 156   & 8     & 41    & 205 \\
          &       & \textbf{Test} & 325   & 353   & 307   & 293   & 19    & 124   & 436 \\
\cmidrule{2-10}          & \multirow{3}[2]{*}{$\mathcal{D}_2$} & \textbf{Train} & 605   & 582   & 462   & 783   & 25    & 205   & 1013 \\
          &       & \textbf{Dev} & 148   & 191   & 183   & 185   & 11    & 53    & 249 \\
          &       & \textbf{Test} & 322   & 347   & 310   & 317   & 25    & 143   & 485 \\
    \midrule
    \multirow{6}[4]{*}{\textbf{16Res}} & \multirow{3}[2]{*}{$\mathcal{D}_1$} & \textbf{Train} & 863   & 775   & 602   & 890   & 43    & 280   & 1213 \\
          &       & \textbf{Dev} & 216   & 270   & 237   & 224   & 8     & 66    & 298 \\
          &       & \textbf{Test} & 328   & 342   & 282   & 360   & 25    & 72    & 457 \\
\cmidrule{2-10}          & \multirow{3}[2]{*}{$\mathcal{D}_2$} & \textbf{Train} & 857   & 759   & 623   & 1015  & 50    & 329   & 1394 \\
          &       & \textbf{Dev} & 210   & 251   & 221   & 252   & 11    & 76    & 339 \\
          &       & \textbf{Test} & 326   & 338   & 282   & 407   & 29    & 78    & 514 \\
    \bottomrule
    \hline
    \end{tabular}%
    
    }
  \caption{Statistic information of our two experiment datasets: 
  ``\#S'', ``\#T'', ``\#A'', and ``\#O'' denote the numbers of ``Sentences'', ``Triplets'', ``Aspects'', and ``Opinions''; ``\#S1'', ``\#S2'', \#S3'' denote the numbers of sentiments ``Positive'', ``Neutral'' and ``Negative'', respectively.}
  \label{tab:data}%
\end{table}%


\subsection{Rethinking the GTS}
\label{Rethinking the GTS}
Rethinking the 2D tagging scheme:

\noindent \textbf{Lemma 1}. Specific to the ASTE task, when we take it as a 2D-labeling problem, we are to 1) find a set of tagging strategies to establish a 1-1 map between each triplet and its corresponding tagging matrix. See the proof in \textbf{Appendix} Proof 1. 
\label{lemma1}

\noindent \textbf{Lemma 2}. In a 2D-tagging for ASTE, at least three basic goals must be met: 
1) correctly identifying the \texttt{(Aspect, Opinion)} pairs, 
2) correctly classifying the sentiment polarity of the pair based on the context, and 
3) avoiding boundary errors, such as \textit{overlapping}\footnote{It occurs when one single word belongs to multiple classes in different triplets. }, \textit{confusion}\footnote{It occurs when there is a lack of location restrictions so that multiple neighbored candidates can not be uniquely distinguished. }, 
and \textit{conflict}\footnote{It occurs when one single word is composed of multiple tokens, and the predict gives predictions that are not aligned with the word span. }. See the proof in \textbf{Appendix} Proof 2.


\noindent \textbf{Theorem 1}. From insight of the above lemmas, it can be concluded that using \textbf{enough} (that is, following the 1-1 map properties in Lemma 1, as well as avoiding the issues in Lemma 2) labels will make it a theoretically ensured tagging scheme.

\noindent \textbf{Assumption 1}. Ceteris paribus, for a specific classification neural network, the \textbf{fewer} the number of target categories, the easier it is for the network to learn. This is a empirical and heuristic assumption, for the reasonable consideration of \textit{Simplification of Decision Boundaries} \cite{Hinton2006} and \textit{Enhancement of Training Efficiency} (less parameters). 

Combining Theorem 1 and Assumption 1, \textbf{fewer} yet \textbf{enough} labels can be heuristically better solution with theoretical guarantee.   



\begin{figure*}[!ht]
\begin{center}
\includegraphics[scale=0.38]{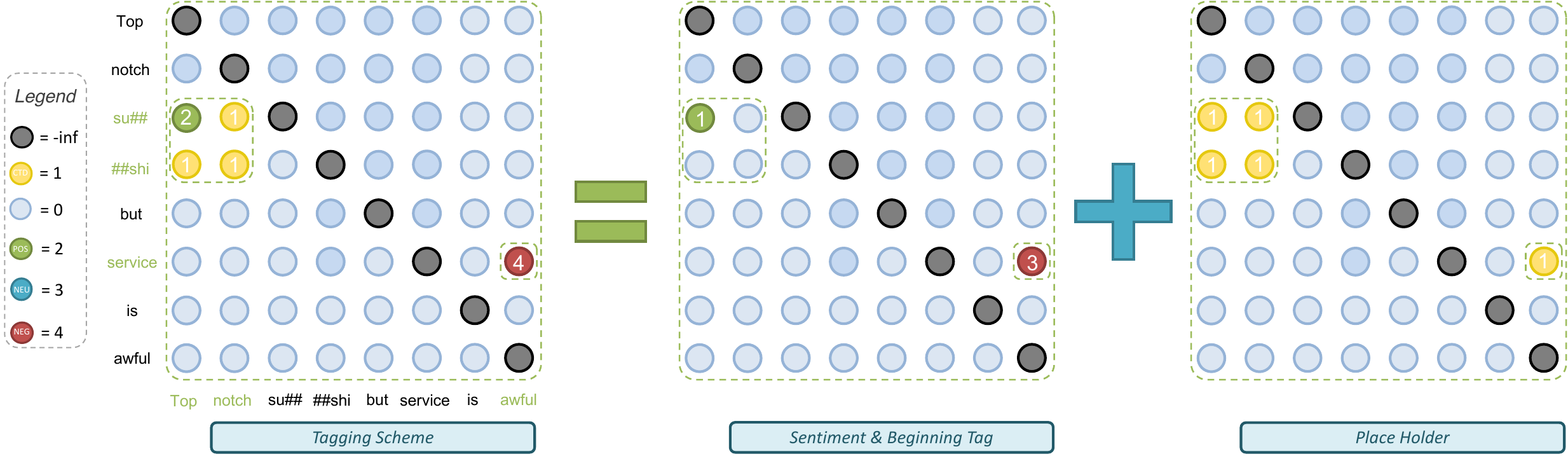} 
\caption{Decomposition of the tagging scheme into two components: 1) a beginning mark matrix with sentiment labels; and 2) a placeholder matrix denoting regions of triplets with ``1''s and default regions with ``0''s. Remember that each row is taken as candidates for an \texttt{Aspect} and each column is taken as candidates for an \texttt{Opinion}. Naturally, each cell in the square matrix can be seen as an ordered pair for a unique candidate of \texttt{<Aspect, Opinion>}. When we simply sum the two components up, we have the left-hand tagging scheme in Figure \ref{fig.tagging}, where the ``Sentiment \& Beginning Tag'' is like a trigger (just like you click your mouse), and the ``Place Holder'' is like a ``continued shift'' (continue to hold and drag the mouse to the downright).}
\label{fig.tagging}
\end{center}
\end{figure*}

With the above knowledge, our tagging scheme employs a full matrix (illustrated as Figure \ref{fig.tagging}) so that rectangular occupations in its cells indicate respective triplets, where each of the rectangles' row indices correspond to the relative \texttt{Aspect} term and the column indices correspond to the \texttt{Opinion}. Hereafter, this kind of labels can be taken as a set of ``place holder'', which is obviously a 1-1 map meeting Lemma 1. 


To further satisfy Lemma 2, we introduce another kind of labels, ``sentiment \& beginning tag''. This set of labels specializes in recognizing the top-left corner of a ``shadowed'' area. Meanwhile, it takes a value from the sentiment polarity, i.e. \texttt{Positive, Neutral, Negative}. This tagging is crucial to both \textit{identify the beginning of an triplet} and \textit{label the sentiment polarity}. 

Figure \ref{fig.tagging} shows a comprehensive case of our tagging scheme, in which the left matrix is an appearance of our tagging scheme, and it can be decomposed into two separate components. The middle matrix is the first component,  which takes only one tag to locate the up-left beginning of an area, and the second component simply predicts a binary classification to figure out the full area.

Note that, this design benefits the tagging scheme's decode process. By scanning across the matrix, we only start an examination function when triggered by a beginning label like this, and then search by row and column until it meets any label except a ``continued'' (``CTD''), which satisfies Lemma 2.

\subsection{Proof 1: }
\begin{adjustwidth}{0.9em}{0pt}
Let:
\begin{itemize}
    \item \( S \) be a sentence with \( n \) tokens.
    \item \( M \) be an \( n \times n \) tagging matrix for \( S \), where each entry \( M[i][j] \) can hold a label.
    \item \( T_k = (A_k, O_k, S_k) \) be a sentiment triplet consisting of an aspect term \( A_k \), an opinion term \( O_k \), and a sentiment \( S_k \).
\end{itemize}

\paragraph{Tagging Strategy}
If \( A_k \) starts at position \( i \) and \( O_k \) starts at position \( j \), then \( M[i][j] \) is tagged with a unique label \( L_k \) that encodes \( S_k \). This label \( L_k \) uniquely identifies the triplet \( T_k \), ensuring that no other entry \( M[i'][j'] \) with \( (i', j') \neq (i, j) \) carries the same label unless it refers to the same sentiment context.

\[
\text{Define } L_k = \text{"start of triplet"} T_k \text{ with sentiment } S_k
\]

\paragraph{Proof of One-to-One Mapping}
\begin{itemize}
    \item \textbf{Injectivity}: Each \( L_k \) uniquely identifies a triplet \( T_k \). If \( M[i][j] = M[i'][j'] = L_k \), then by definition, \( (i, j) = (i', j') \) and \( T_k \) is the same.
    \item \textbf{Surjectivity}: Each triplet \( T_k \) can be uniquely located and identified by its label \( L_k \) in matrix \( M \), where no two distinct triplets have the same label at the same matrix position.
\end{itemize}

\paragraph{Conclusion}
The tagging scheme ensures that each sentiment triplet \( T_k \) is uniquely mapped to a specific label in the matrix \( M \), and each label in \( M \) uniquely refers back to a specific triplet \( T_k \). This guarantees a one-to-one correspondence between the triplets and their tagging matrix representations, fulfilling the conditions required by Lemma 1 for an effective and efficient ASTE process.

\end{adjustwidth}

\subsection{Proof 2: }
\label{Proof 2}
\begin{adjustwidth}{0.9em}{0pt}
For the ASTE task, considered as a 2D-labeling problem, it is necessary to ensure three fundamental goals are met:

\paragraph{Definitions}
\begin{itemize}
    \item \( S \) be a sentence with \( n \) tokens.
    \item \( M \) be an \( n \times n \) tagging matrix for \( S \), where each entry \( M[i][j] \) can hold a label indicating a component of a sentiment triplet.
    \item \( T_k = (A_k, O_k, S_k) \) be a sentiment triplet consisting of an aspect term \( A_k \), an opinion term \( O_k \), and a sentiment \( S_k \).
\end{itemize}

\paragraph{Goals}
\begin{enumerate}
    \item \textbf{Correct Identification of Pairs}: Ensure that each (Aspect, Opinion) pair is correctly identified in the tagging matrix \( M \).
    \item \textbf{Classification of Sentiment Polarity}: Accurately classify the sentiment polarity \( S_k \) for each (Aspect, Opinion) pair.
    \item \textbf{Avoidance of Boundary Errors}: Prevent boundary errors such as overlapping and confusion in the tagging matrix \( M \).
\end{enumerate}

\paragraph{Proof Using Contraposition}
\begin{enumerate}
    \item \textbf{Assuming Incorrect Identification}:
    Assume that some (Aspect, Opinion) pairs are incorrectly identified in \( M \). This would mean that there exists at least one pair \( (i, j) \) where \( M[i][j] \) does not represent the actual (Aspect, Opinion) relationship in \( S \). This misrepresentation leads to incorrect sentiment analysis results, which contradicts the requirement of the task to provide accurate sentiment analysis, thereby proving that our identification must be correct.

    \item \textbf{Assuming Incorrect Classification}:
    Assume the sentiment polarity \( S_k \) is incorrectly classified in \( M \). This would imply that the sentiment associated with an (Aspect, Opinion) pair is wrong, leading to a sentiment analysis that does not reflect the true sentiment of the text. Given that the primary goal of ASTE is to accurately identify sentiments, this assumption leads to a contradiction, thereby establishing that our classification must be accurate.

    \item \textbf{Assuming Existence of Boundary Errors}:
    Assume boundary errors such as overlaps or confusion occur in \( M \). Such errors would prevent the clear identification and classification of sentiment triplets, leading to incorrect or ambiguous extraction outcomes. This would undermine the integrity and usability of the ASTE process, contradicting the task's need for precise extraction mechanisms. Hence, we prove that boundary errors must be effectively managed.
\end{enumerate}

\paragraph{Conclusion}
The contraposition approach solidifies that the tagging strategy for ASTE in a 2D labeling framework successfully achieves the correct identification of pairs, accurate classification of sentiment, and effective management of boundary errors, as any failure in these aspects leads to contradictions with the task requirements.

\end{adjustwidth}

\subsection{Baselines}
\label{sec:Baseline}
See Table \ref{tab:baselines}. 

\begin{table*}
\centering
  \scalebox{0.65}{
    \begin{tabular}{lp{40em}}
    \hline
    \toprule
    \textbf{Methods} &  \textbf{Brief Introduction}  \\
    \midrule
    \midrule
    \textbf{Pipeline} &    \\
    OTE-MTL \textrm{\cite{zhang2020multi}} & It proposes a multi-task learning framework including two parts: aspect and opinion tagging, along with word-level sentiment dependency parsing. This approach simultaneously extracts aspect and opinion terms while parsing sentiment dependencies using a biaffine scorer. Additionally, it employs triplet decoding based on the aforementioned outputs during inference to facilitate triplet extraction. \\
    \textrm{Li-unified-R+PD}\ \textrm{\cite{peng2020knowing}} &   It proposes an unified tagging scheme, Li-unified-R, to assist target boundary detection. Two stacked LSTMs are employed to complete aspect-based sentiment prediction and the sequence labeling. \\
    \textrm{CMLA+C-GCN}\ \textrm{\cite{wang2017coupled}} & It facilitates triplet extraction by modelling the interaction between the aspects and opinions.  \\
    Two-satge\ \textrm{\cite{peng2020knowing}} & It decomposes triplet extraction to two stages: 1)  predicting unified aspect-sentiment and opinion tags; and 2) pairing the two results from stage one. \\
    \textrm{RI-NANTE+} \textrm{\cite{dai2019neural}} & It adopts the same sentiment triplets extracting method as that of \textrm{CMLA+}, but it incorporates a novel LSTM-CRF mechanism and fusion rules to capture word dependencies within sentences.\\

    \midrule
    \textbf{Sequence-tagging} &       \\
    Span-BART \textrm{\cite{yan2021unified}} & It redefines triplet extraction within an end-to-end framework by utilizing a sequence composed of pointer and sentiment class indexes. This is achieved by leveraging the pretrained sequence-to-sequence model BART to address ASTE. \\
    JET \textrm{\cite{xu2020position}}   &  It extracts triplets jointly by designing a position-aware sequence-tagging scheme to extract the triplets and capturing the rich interactions among the elements.  \\
    \midrule
    \textbf{Seq2seq} &    \\
    Dual-MRC \textrm{\cite{mao2021joint}} & It proposes a solution for ASTE by jointly training two BERT-MRC models with parameters sharing. \\
    BMRC \textrm{\cite{chen2021bidirectional}} & It introduces a bidirectional MRC (BMRC) framework for ASTE, employing three query types: non-restrictive extraction queries, restrictive extraction queries, and sentiment classification queries. The framework synergistically leverages two directions, one for sequential recognition of aspect-opinion-sentiment and the other for sequential recognition of opinion-aspects-sentiment expressions. \\
    \midrule
    \textbf{Table-filling} &   \\
    GTS \textrm{\cite{wu2020grid}}   & It proposes a novel 2D tagging scheme to address ASTE in an end-to-end fashion only with one unified grid tagging task. 
    It also devises an effective inference strategy on GTS that utilizes mutual indication between different opinion factors to achieve more accurate extraction. \\
    Double-encoder \textrm{\cite{jing2021seeking}} & It proposes a dual-encoder model that capitalizes on encoder sharing while emphasizing differences to enhance effectiveness. 
    One of the encoders, referred to as the pair encoder, specifically concentrates on candidate aspect-opinion pair classification, while the original encoder retains its focus on sequence labeling. \\
    $\mathrm{S}^3\mathrm{E}^2$ \textrm{\cite{chen2021semantic}} &  It represents the semantic and syntactic relationships between word pairs, employs GNNs for encoding, and applies a more efficient inference strategy. \\
    EMC-GCN \textrm{\cite{chen2022enhanced}} & It employs a biaffine attention module to embed ten types of relations within sentences, transforming the sentence into a multi-channel graph while incorporating various enhanced linguistic features to enhance performance. 
    Additionally, the method introduces an effective strategy for refining word-pair representations, aiding in the determination of whether word pairs are a match or not.\\
    \midrule
    \textbf{LLM-based} &    \\
    zero-shot & Performing aspect-based sentiment analysis using an LLM. The specific method involves inputting a prompted sentence and directly outputting the corresponding [A, O, S] triplets. An example of the text given to the LLM, with the prompt added, is as follows: \textcolor{cyan}{"Perform aspect-based sentiment analysis on the provided text and return triplets as [Aspect, Opinion, Sentiment]. You only need to provide the triplets, no additional explanations are required. The provided text: \{sentence\}"}\\ 

    few-shot & Building upon the zero-shot method, a small number of examples from the training set are added to the prompted sentence: \textcolor{cyan}{"Perform aspect-based sentiment analysis on the provided text and return triplets as [Aspect, Opinion, Sentiment]. For example: input: \{train sentence\} output: \{train triplets\}, ... (some other examples). You only need to provide the triplets, no additional explanations are required. The provided text: \{sentence\}"}. We utilized 5-shot, 10-shot, and 20-shot methods, all randomly sampled from the training set. The results indicate that the 5-shot method performed the best, while the performances of the 10-shot and 20-shot methods showed a decline. The tables presents the output results for the 5-shot method.\\
    \bottomrule
    \hline
    \end{tabular}%
    }
\caption{Baseline methods with brief introduction.}
\label{tab:baselines}
\end{table*}

\subsection{Performance on Other ABSA Tasks}
\label{sec:Performance on Other ABSA Tasks}
See Table \ref{tab:otherABSAperformance}. 

\begin{table*}[htbp]
  \centering
  \scalebox{0.68}{
    \begin{tabular}{cccccccccccccccc}
    \hline
    \toprule
    \multirow{2}[4]{*}{\textbf{Methods}} & \multicolumn{3}{c}{\textbf{14Res}} &       & \multicolumn{3}{c}{\textbf{14Lap}} &       & \multicolumn{3}{c}{\textbf{15Res}} &       & \multicolumn{3}{c}{\textbf{16Res}} \\
\cmidrule{2-4}\cmidrule{6-8}\cmidrule{10-12}\cmidrule{14-16}          & \textbf{AE} & \textbf{OE} & \textbf{AOPE} &       & \textbf{AE} & \textbf{OE} & \textbf{AOPE} &       & \textbf{AE} & \textbf{OE} & \textbf{AOPE} &       & \textbf{AE} & \textbf{OE} & \textbf{AOPE} \\
    \midrule
    \midrule
    CMLA  & 81.22 & 83.07 & 48.95 &       & 78.68 & 77.95 & 44.10 &       & 76.03 & 74.67 & 44.60 &       & 74.20 & 72.20 & 50.00 \\
    RINANTE & 81.34 & 83.33 & 46.29 &       & 77.13 & 75.34 & 29.70 &       & 73.38 & 75.40 & 35.40 &       & 72.82 & 70.45 & 30.70 \\
    Li-unified & 81.62 & 85.26 & 55.34 &       & 78.54 & 77.55 & 52.56 &       & 74.65 & 74.25 & 56.85 &       & 73.36 & 73.87 & 53.75 \\
    GTS   & 83.82 & 85.04 & 75.53 &       & 79.52 & 78.61 & 65.67 &       & 78.22 & 79.31 & 67.53 &       & 75.80 & 76.38 & 74.62 \\
    Dual-MRC & \textbf{86.60} & 86.22 & 77.68 &       & 80.44 & 79.90 & 63.37 &       & 75.08 & 77.52 & 64.97 &       & 76.87 & 77.90 & 75.71 \\
    \midrule
    \textbf{MiniConGTS (Ours)} & 86.55 & \textbf{87.04} & \textbf{79.60} &       & \textbf{82.62} & \textbf{83.41} & \textbf{73.23} &       & \textbf{86.53} & \textbf{83.05} & \textbf{73.87} &       & \textbf{85.48} & \textbf{87.06} & \textbf{76.29} \\
    $\Delta$F1     & -0.05 & 0.82  & 1.92  &       & 2.18  & 3.51  & 7.56  &       & 8.31 & 3.74  & 6.34  &       & 8.61  & 9.16  & 0.58 \\

    \bottomrule
    \hline
    \end{tabular}%
  }
  \caption{F1-score performance on other ABSA tasks: AE, OE, and AOPE. The test is implemented on $\mathcal{D}_1$. Results of other models are retrieved from \cite{fei2022inheriting}. }
  \label{tab:otherABSAperformance}%
\end{table*}%

\subsection{ASTE Performance on ($\mathcal{D}_1$) }
\label{sec:ASTE Performance on D1}
See Table \ref{tab:D1performance}.

\begin{table*}[ht]
\centering
  \scalebox{0.68}{
    \begin{tabular}{cccccccccccccccc}
    \hline
    \toprule
    \multirow{2}[4]{*}{\textbf{Methods}} & \multicolumn{3}{c}{\textbf{14Res}} &       & \multicolumn{3}{c}{\textbf{14Lap}} &       & \multicolumn{3}{c}{\textbf{15Res}} &       & \multicolumn{3}{c}{\textbf{16Res}} \\
\cmidrule{2-4}\cmidrule{6-8}\cmidrule{10-12}\cmidrule{14-16}          & \multicolumn{1}{c}{P} & R     & F1    &       & \multicolumn{1}{c}{P} & \multicolumn{1}{c}{R} & \multicolumn{1}{c}{F1} &       & \multicolumn{1}{c}{P} & R     & F1    &       & \multicolumn{1}{c}{P} & \multicolumn{1}{c}{R} & \multicolumn{1}{c}{F1} \\
    \midrule
    \midrule
    \textbf{Pipeline} &       &       &       &       &       &       &       &       &       &       &       &       &       &       &  \\
    OTE-MTL \textrm{\cite{zhang2020multi}} & -     & -     & 45.05 &       & -     & -     & 59.67 &       & -     & -     & 48.97 &       & -     & -     & 55.83 \\
    $\textrm{Li-unified-R+PD}^\sharp$\ \textrm{\cite{peng2020knowing}} & 41.44     & 68.79     & 51.68 &       & 42.25     & 42.78     & 42.47 &       & 43.34     & 50.73     & 46.69 &       & 38.19     & 53.47     & 44.51 \\
    \textrm{RI-NANTE+} \textrm{\cite{dai2019neural}} & 31.42 & 39.38 & 34.95 &       & 21.71 & 18.66 & 20.07 &       & 29.88 & 30.06 & 29.97 &       & 25.68 & 22.30 & 23.87 \\
    $\textrm{CMLA+C-GCN}^\flat$\ \textrm{\cite{wang2017coupled}} & 72.22 & 56.35 & 63.17 &       & 60.69 & 47.25 & 53.03 &       & 64.31 & 49.41 & 55.76 &       & 66.61 & 59.23 & 62.70 \\
    Two-satge$^\natural$\ \textrm{\cite{peng2020knowing}} & 58.89 & 60.41 & 59.64 &       & 48.62 & 45.52 & 47.02 &       & 51.7  & 46.04 & 48.71 &       & 59.25 & 58.09 & 59.67 \\
    \midrule
    \textbf{Sequence-tagging} &       &       &       &       &       &       &       &       &       &       &       &       &       &       &  \\
    Span-BART \textrm{\cite{yan2021unified}} & -     & -     & 72.46 &       & -     & -     & 57.59 &       & -     & -     & 60.10 &       & -     & -     & 69.98 \\
    JET \textrm{\cite{xu2020position}}   & 67.97 & 60.32 & 63.92 &       & 58.47 & 43.67 & 50.00 &       & 58.35 & 51.43 & 54.67 &       & 64.77 & 61.29 & 62.98 \\
    \midrule
    \textbf{MRC based} &       &       &       &       &       &       &       &       &       &       &       &       &       &       &  \\
    BMRC$^\dagger$\ \textrm{\cite{chen2021bidirectional}}  & 71.32 & 70.09 & 70.69 &       & 65.12 & 54.41 & 59.27 &       & 63.71 & 58.63 & 61.05 &       & 67.74 & 68.56 & 68.13 \\
    COM-MRC \textrm{\cite{zhai2022mrc}} & \underline{76.45} & 69.67 & 72.89 &       & 64.73 & 56.09 & 60.09 &       & 68.50 & 59.74 & 63.65 &       & \underline{72.80} & 70.85 & 71.79 \\
    \midrule
    \textbf{Table-filling} &       &       &       &       &       &       &       &       &       &       &       &       &       &       &  \\
    $\mathrm{S}^3\mathrm{E}^2$ \textrm{\cite{chen2021semantic}} & 69.08 & 64.55 & 66.74 &       & 59.43 & 46.23 & 52.01 &       & 61.06 & 56.44 & 58.66 &       & 71.08 & 63.13 & 66.87 \\
    GTS \textrm{\cite{wu2020grid}}   & 70.92 & 69.49 & 70.20 &       & 57.52 & 51.92 & 54.58 &       & 59.29 & 58.07 & 58.67 &       & 68.58 & 66.60 & 67.58 \\
    EMC-GCN \textrm{\cite{chen2022enhanced}} & 71.85 & 72.12 & 71.78 &       & 61.46 & \underline{55.56} & 58.32 &       & 59.89 & 61.05 & 60.38 &       & 65.08 & 71.66 & 68.18 \\
    BDTF \textrm{\cite{zhang2022boundary}} & \textbf{76.71} & \underline{74.01} & \underline{75.33} &       & \textbf{68.30}  & 55.10  & \underline{60.99} &       & \textbf{66.95} & \textbf{65.05} & \textbf{65.97} &       & \textbf{73.43} & \underline{73.64} & \textbf{73.51} \\
    DGCNAP \textrm{\cite{li2023dual}} & 71.83 & 68.77 & 70.26 &       & 66.46 & 54.34 & 58.74 &       & 62.03 & 57.18  & 59.49 &       & 69.39 & 72.20 & 70.77 \\
    \midrule

    \textbf{LLM-based} &       &       &       &       &       &       &       &       &       &       &       &       &       &       &  \\
    GPT-3.5-turbo zero-shot  & 39.21     & 56.17     & 46.18 &       & 26.21     & 40.69     & 31.88 &       & 31.21     & 52.75     & 39.21 &       & 35.28     & 59.64    & 44.34 \\
    $\textrm{GPT-3.5-turbo few-shot}$\  & 50.32     & 64.75     & 56.63 &       & 29.67     & 43.90     & 35.41 &       & 36.94     & 61.01     & 46.02 &       & 44.80     & 69.96     & 54.62 \\
    $\textrm{GPT-3.5-turbo CoT}$\  & 40.78     & 57.93     & 47.86 &       & 28.37     & 43.25     & 34.27 &       & 35.17     & 57.11     & 43.53 &       & 40.32     & 65.79     & 50.00 \\
    $\textrm{GPT-3.5-turbo CoT+few-shot}$\  & 44.97     & 57.81     & 50.59 &       & 28.31     & 43.04     & 34.15 &       & 35.71     & 58.49     & 44.35 &       & 43.72     & 66.45     & 52.74 \\
    \midrule
    
    \textbf{Ours} &       &       &       &       &       &       &       &       &       &       &       &       &       &       &  \\
      MiniConGTS    & 75.87 & \textbf{76.12} & \textbf{76.00} &       & \underline{67.45} & \textbf{61.01} & \textbf{64.07} &       & \underline{66.84} & \underline{64.08} & \underline{65.43} &       & 69.38 & \textbf{74.40} & \underline{71.80} \\
    \bottomrule
    \hline
    \end{tabular}
}
    \caption{Experimental results on $\mathcal{D}_1$ \cite{wu2020grid}. The best results are highlighted in bold,  while the second best results are underscored. The results with $\dagger$ are retrieved from \cite{yu2021aspect}. The results with $\natural$ are retrieved from \cite{xu2020position}. 
The results with $\flat$ are retrieved from \cite{wu2020grid}. 
The results with $\sharp$ are retrieved from \cite{peng2020knowing}.
}
\label{tab:D1performance}
\end{table*}

\subsection{Details Settings of GPT Experients}
\label{sec:Details of GPT Experients}
See Table \ref{tab:gptExample}. 

\begin{sidewaystable*}[htbp]
\centering
\tiny
\renewcommand{\arraystretch}{1.5}
\rotatebox{0}{
\begin{tabular}{p{5cm}@{\hskip 0.5cm}p{3cm}@{\hskip 0.5cm}p{3cm}@{\hskip 0.5cm}p{3cm}@{\hskip 0.5cm}p{3cm}@{\hskip 0.5cm}p{3cm}}
\toprule
\multirow{2}{*}{Sentence} & \multirow{2}{*}{Ground Truth} & \multicolumn{4}{c}{Predictions} \\ \cline{3-6} 

& & GPT-3.5 zero-shot & GPT-3.5 few-shot & GPT-4 zero-shot & GPT-4 few-shot \\ 
\midrule
\midrule
It is a cozy place to go with a couple of friends. & [place, cozy, positive] & [place, cozy, positive],\newline [friends, couple of, positive] & [place, cozy, positive] & [place, cozy, positive]\newline [going with friends, a couple of, positive] & [place, cozy, positive] \\ \midrule
This place has the best sushi in the city. & [sushi, best, positive] & [sushi, best, positive],\newline [place, best, positive], [city, best, positive] & [sushi, best, positive] & [sushi, the best, positive],\newline [place, This, neutral],\newline [city, in the, neutral] & [sushi, best, positive] \\ \midrule
Disappointingly, their wonderful Saketini has been taken off the bar menu. & [Saketini, wonderful, positive],\newline [bar menu, Disappointingly, negative] & [Saketini, wonderful, positive],\newline [bar menu, taken off, negative] & [Saketini, taken off, negative] & [Saketini, has been taken off the bar menu, negative],\newline [bar menu, wonderful, positive] & [Saketini, wonderful, positive],\newline [bar menu, taken off, negative] \\ \midrule
I went in one day asking for a table for a group and was greeted by a very rude hostess. & [hostess, rude, negative] & [hostess, rude, negative] & [hostess, rude, negative] & [table for a group, asking, neutral],\newline [hostess, very rude, negative] & [table for a group, asking, neutral],\newline [hostess, very rude, negative] \\ \midrule
But make sure you have enough room on your credit card as the bill will leave a big dent in your wallet. & [bill, big, negative] & [credit card, enough room, positive],\newline [bill, big dent, negative],\newline [wallet, big dent, negative] & [credit card, enough room, negative],\newline [bill, big dent, negative],\newline [wallet, big dent, negative] & [room on your credit card, enough, positive],\newline [bill, will leave a big dent in your wallet, negative] & [bill, big dent, negative] \\ \bottomrule
\end{tabular}
}

\caption{In summary, there are several challenges observed in the performance of GPT models concerning triplets. Firstly, there is a prominent issue of "hard" matching, where GPT models tend to introduce additional modifiers or adverbs in the opinion component, leading to a lack of exact correspondence. Secondly, during zero-shot inference, GPT models tend to generate multiple predicted triplets, resulting in decreased precision. This behavior particularly hampers the precision of the model's predictions. Thirdly, inconsistencies arise in handling triplets involving structures such as [A, O1 and O2, S] and [A, O1, S], [A, O2, S]. This inconsistency is challenging to mitigate due to its dependence on annotation practices and conventions.
Upon closer examination, the issues observed do not appear to be as pronounced as indicated by the evaluation metrics. Rather, they often manifest as cases where the general idea is correctly captured, but the precise format or phrasing does not align perfectly. Notably, the performance of GPT-4 deteriorates due to its occasional tendency to not merely "extract" fragments from sentences but to generate its own summarizations. Consequently, evaluating against triplets that originate solely from annotated sentences poses a challenge in achieving alignment. Furthermore, GPT-4 exhibits a proclivity for extracting longer sequences of words as aspects or opinions, while GPT-3.5 tends to produce shorter sequences that better conform to typical annotation scenarios.}
\label{tab:gptExample}
\end{sidewaystable*}

\begin{table*}
\centering
  \scalebox{0.65}{
    \begin{tabular}{lp{40em}}
    \hline
    \toprule
    \textbf{Methods} &  \textbf{Prompts}  \\
    \midrule
    \midrule

    \textbf{zero-shot} & Suppose you are an expert of aspect-based sentiment analysis. Perform aspect-based sentiment analysis on the provided text and return triplets as [Aspect, Opinion, Sentiment]. You only need to provide the triplets, no additional explanations are required. The provided text: \{sentence\}\\ 
    \midrule
    \textbf{few-shot} & Suppose you are an expert of aspect-based sentiment analysis. Perform aspect-based sentiment analysis on the provided text and return triplets as [Aspect, Opinion, Sentiment]. For example:\\
    &input: The food is uniformly exceptional , with a very capable kitchen which will proudly whip up whatever you feel like eating , whether it 's on the menu or not . \\
    &output: ['food', 'exceptional', 'positive'], ['kitchen', 'capable', 'positive'] \\
    &... \\
    &\textcolor{cyan}{(generated from training set)}\\
    &Now I will provide a new sentence, and you only need to provide the triplets [Aspect, Opinion, Sentiment] without any additional explanations. The provided sentence: \{sentence\}\\
    \midrule
    \textbf{CoT} & Suppose you are an expert of aspect-based sentiment analysis. Please analyze the given text for aspect-based sentiment analysis using the following steps:\\    
    &Definitions:\\
    &- Aspect: An aspect is a specific part or feature of the entity being discussed. It is usually a noun or a noun phrase.\\
    &- Opinion: An opinion is a descriptive term or phrase that expresses a sentiment towards the aspect. It is usually an adjective or a descriptive phrase.\\
    &- Sentiment: The sentiment is the overall feeling expressed towards the aspect, categorized as positive, negative, or neutral.\\
    
    &Instructions:\\
    &1. Read the text and identify all aspects mentioned.\\
    &2. For each identified aspect, determine the opinion expressed and the sentiment (positive, negative, neutral).\\
    &3. Summarize the findings in the format [Aspect, Opinion, Sentiment]. Each triplet must contain an aspect, an opinion, and a sentiment.\\
    &4. If there is a one-to-many relationship between aspects and opinions, list multiple triplets.\\
    &5. Use all words from the original text to answer without any changes.\\
    
    &Example: \textcolor{cyan}{(automatically generated by ChatGPT-4o)}\\
    &Text: "The restaurant has a great ambiance, but the service is poor and the food is average."\\
    
    &Steps:\\
    &1. Identify aspects: ambiance, service, food.\\
    &2. Evaluate opinions and sentiments:\\
    &    - ambiance: Opinion - great, Sentiment - positive\\
    &    - service: Opinion - poor, Sentiment - negative\\
    &    - food: Opinion - average, Sentiment - neutral\\
    &3. Summarize:\\
    &    - [ambiance, great, positive]\\
    &    - [service, poor, negative]\\
    &    - [food, average, neutral]\\
    
    &Please analyze the following text: \{sentence\}\\
    \midrule
    \textbf{CoT+few-shot} & Suppose you are an expert of aspect-based sentiment analysis. Please analyze the given text for aspect-based sentiment analysis using the following steps:\\    
    &Definitions:\\
    &- Aspect: An aspect is a specific part or feature of the entity being discussed. It is usually a noun or a noun phrase.\\
    &- Opinion: An opinion is a descriptive term or phrase that expresses a sentiment towards the aspect. It is usually an adjective or a descriptive phrase.\\
    &- Sentiment: The sentiment is the overall feeling expressed towards the aspect, categorized as positive, negative, or neutral.\\
    
    &Instructions:\\
    &1. Read the text and identify all aspects mentioned.\\
    &2. For each identified aspect, determine the opinion expressed and the sentiment (positive, negative, neutral).\\
    &3. Summarize the findings in the format [Aspect, Opinion, Sentiment]. Each triplet must contain an aspect, an opinion, and a sentiment.\\
    &4. If there is a one-to-many relationship between aspects and opinions, list multiple triplets.\\
    &5. Use all words from the original text to answer without any changes.\\
    
    &Example: \textcolor{cyan}{(generated from training set)}\\
    &Text: ...\\
    
    &Steps:\\
    &1. Identify aspects: ...\\
    &2. Evaluate opinions and sentiments:\\
    &    - ...\\
    &3. Summarize:\\
    &    - [..., ..., ...]\\
    &...\\
    &Please analyze the following text: \{sentence\}\\
    \bottomrule
    \hline
    \end{tabular}%
    }
\caption{LLM prompts in different methods.}
\label{LLM prompts}
\end{table*}

\subsection{Detailed Results of GPT Experiments}
\label{sec:Detailed Results of GPT Experiments}
See Table \ref{tab:gpt-results}. 
\begin{table*}[ht]
\centering
  \scalebox{0.66}{
    \begin{tabular}{cccccccccccccccccccc}
    \hline
    \toprule
    \multirow{2}[4]{*}{\textbf{Method}} & 
    \multirow{2}[4]{*}{\textbf{Combination}} & \multicolumn{3}{c}{\textbf{14Res}} &       & \multicolumn{3}{c}{\textbf{14Lap}} &       & \multicolumn{3}{c}{\textbf{15Res}} &       & \multicolumn{3}{c}{\textbf{16Res}} \\
\cmidrule{3-5}\cmidrule{7-9}\cmidrule{11-13}\cmidrule{15-17}       &   & \multicolumn{1}{c}{P} & R     & F1    &       & \multicolumn{1}{c}{P} & \multicolumn{1}{c}{R} & \multicolumn{1}{c}{F1} &       & \multicolumn{1}{c}{P} & R     & F1    &       & \multicolumn{1}{c}{P} & \multicolumn{1}{c}{R} & \multicolumn{1}{c}{F1} \\
    \midrule
    \midrule
          & A-O-S & 0.5151 & 0.6519 & 0.5755 &       & 0.3979 & 0.5009 & 0.4435 &       & 0.4334 & 0.6309 & 0.5139 &       & 0.5112 & 0.7101 & 0.5945 \\
          & A-O   & 0.5429 & 0.6871 & 0.6066 &       & 0.4479 & 0.5638 & 0.4992 &       & 0.4788 & 0.6969 & 0.5676 &       & 0.5420 & 0.7529 & 0.6303 \\
          & A-S   & 0.6234 & 0.7983 & 0.7001 &       & 0.4955 & 0.7149 & 0.5853 &       & 0.5539 & 0.7731 & 0.6454 &       & 0.5853 & 0.8274 & 0.6856 \\
          GPT-3.5-turbo & O-S   & 0.5790 & 0.7438 & 0.6512 &       & 0.5016 & 0.6653 & 0.5719 &       & 0.5228 & 0.7281 & 0.6086 &       & 0.6020 & 0.7851 & 0.6814 \\
          few-shot & A     & 0.6758 & 0.8455 & 0.7512 &       & 0.5805 & 0.8337 & 0.6844 &       & 0.6220 & 0.838 & 0.7140 &       & 0.6361 & 0.8739 & 0.7363 \\
          & O     & 0.6138 & 0.7938 & 0.6923 &       & 0.5587 & 0.7505 & 0.6406 &       & 0.5780 & 0.8048 & 0.6728 &       & 0.6444 & 0.8404 & 0.7295 \\
          & S     & 0.8251 & 0.9222 & 0.8710 &       & 0.7907 & 0.8844 & 0.8349 &       & 0.8286 & 0.9337 & 0.878 &       & 0.8179 & 0.9466 & 0.8776 \\
    \midrule
          & A-O-S & 0.4847 & 0.5905 & 0.5324 &       & 0.3048 & 0.4030 & 0.3471 &       & 0.3951 & 0.5670 & 0.4657 &       & 0.4403 & 0.6381 & 0.5210 \\
          & A-O   & 0.5178 & 0.6308 & 0.5687 &       & 0.3566 & 0.4713 & 0.4061 &       & 0.4368 & 0.6268 & 0.5148 &       & 0.4711 & 0.6829 & 0.5576 \\
          & A-S   & 0.5991 & 0.7807 & 0.6779 &       & 0.4088 & 0.6199 & 0.4927 &       & 0.5123 & 0.7685 & 0.6148 &       & 0.5199 & 0.8075 & 0.6326 \\
          GPT-3.5-turbo & O-S   & 0.5708 & 0.7143 & 0.6345 &       & 0.4288 & 0.5768 & 0.4919 &       & 0.4834 & 0.6711 & 0.5620 &       & 0.5481 & 0.7511 & 0.6338 \\
          CoT & A     & 0.6594 & 0.8538 & 0.7441 &       & 0.5142 & 0.7797 & 0.6197 &       & 0.5901 & 0.8796 & 0.7063 &       & 0.5702 & 0.8805 & 0.6922 \\
          & O     & 0.6119 & 0.7723 & 0.6828 &       & 0.4930 & 0.6716 & 0.5686 &       & 0.5403 & 0.7500  & 0.6281 &       & 0.5932 & 0.8128 & 0.6858 \\
          & S     & 0.7942 & 0.9374 & 0.8599 &       & 0.7229 & 0.9046 & 0.8036 &       & 0.7477 & 0.9222 & 0.8258 &       & 0.7494 & 0.9585 & 0.8411 \\
    \midrule
          & A-O-S & 0.4941 & 0.5915 & 0.5385 &       & 0.3378 & 0.4233 & 0.3757 &       & 0.3902 & 0.5608 & 0.4602 &       & 0.4649 & 0.6693 & 0.5486 \\
          & A-O   & 0.5294 & 0.6338 & 0.5769 &       & 0.3968 & 0.4972 & 0.4413 &       & 0.4333 & 0.6227 & 0.5110 &       & 0.4932 & 0.7101 & 0.5821 \\
          & A-S   & 0.6257 & 0.7925 & 0.6993 &       & 0.4518 & 0.6479 & 0.5324 &       & 0.5306 & 0.7824 & 0.6324 &       & 0.5451 & 0.8296 & 0.6579 \\
          GPT-3.5-turbo & O-S   & 0.5735 & 0.7048 & 0.6324 &       & 0.4681 & 0.6168 & 0.5322 &       & 0.4936 & 0.6776 & 0.5712 &       & 0.5769 & 0.7745 & 0.6612 \\
          CoT+few-shot& A     & 0.6907 & 0.8691 & 0.7697 &       & 0.5572 & 0.7991 & 0.6566 &       & 0.6026 & 0.8704 & 0.7121 &       & 0.5873 & 0.8783 & 0.7039 \\
          & O     & 0.6186 & 0.7676 & 0.6851 &       & 0.5319 & 0.7100  & 0.6082 &       & 0.5479 & 0.7522 & 0.6340 &       & 0.6149 & 0.8255 & 0.7048 \\
          & S     & 0.8119 & 0.9336 & 0.8685 &       & 0.7470 & 0.8960 & 0.8147 &       & 0.7775 & 0.9366 & 0.8497 &       & 0.7923 & 0.9733 & 0.8735 \\
    \midrule
    & A-O-S & 0.5411 & 0.6620 & 0.5955 &       & 0.3823 & 0.4861 & 0.4280 &       & 0.4557 & 0.6041 & 0.5195 &       & 0.5290 & 0.7101 & 0.6063 \\
          & A-O   & 0.5757 & 0.7042 & 0.6335 &       & 0.439 & 0.5582 & 0.4915 &       & 0.5023 & 0.6660 & 0.5727 &       & 0.5551 & 0.7451 & 0.6362 \\
          & A-S   & 0.6777 & 0.8208 & 0.7424 &       & 0.4962 & 0.7084 & 0.5836 &       & 0.5872 & 0.7639 & 0.6640 &       & 0.6019 & 0.8296 & 0.6977 \\
          GPT-4o & O-S   & 0.6047 & 0.7532 & 0.6709 &       & 0.4863 & 0.6337 & 0.5503 &       & 0.5613 & 0.7325 & 0.6356 &       & 0.6318 & 0.8106 & 0.7102 \\
          few-shot & A     & 0.7367 & 0.8679 & 0.7970 &       & 0.5951 & 0.8445 & 0.6982 &       & 0.6611 & 0.8218 & 0.7327 &       & 0.6471 & 0.8761 & 0.7444 \\
          & O     & 0.6389 & 0.8033 & 0.7117 &       & 0.5517 & 0.7271 & 0.6274 &       & 0.6185 & 0.8070 & 0.7003 &       & 0.6667 & 0.8553 & 0.7493 \\
          & S     & 0.8353 & 0.9336 & 0.8817 &       & 0.7758 & 0.8699 & 0.8202 &       & 0.8376 & 0.9366 & 0.8844 &       & 0.8346 & 0.9585 & 0.8923 \\
    \midrule
          & A-O-S & 0.4121 & 0.5332 & 0.4649 &       & 0.2698 & 0.3771 & 0.3146 &       & 0.3307 & 0.5093 & 0.4010 &       & 0.3914 & 0.5817 & 0.4679 \\
          & A-O   & 0.4331 & 0.5604 & 0.4886 &       & 0.3122 & 0.4362 & 0.3639 &       & 0.3614 & 0.5567 & 0.4383 &       & 0.4162 & 0.6187 & 0.4977 \\
          & A-S   & 0.6163 & 0.8278 & 0.7066 &       & 0.4486 & 0.6976 & 0.5461 &       & 0.5200  & 0.8125 & 0.6341 &       & 0.5374 & 0.8429 & 0.6563 \\
          GPT-4o & O-S   & 0.4711 & 0.6246 & 0.5371 &       & 0.3439 & 0.4779 & 0.4000   &       & 0.4093 & 0.5987 & 0.4862 &       & 0.4752 & 0.6532 & 0.5502 \\
          CoT   & A     & 0.6667 & 0.8703 & 0.7550 &       & 0.5280 & 0.8143 & 0.6406 &       & 0.5762 & 0.8750 & 0.6949 &       & 0.5799 & 0.8916 & 0.7027 \\
          & O     & 0.5004 & 0.6687 & 0.5724 &       & 0.3951 & 0.5544 & 0.4614 &       & 0.4558 & 0.6667 & 0.5414 &       & 0.5077 & 0.6979 & 0.5878 \\
          & S     & 0.7859 & 0.9545 & 0.8620 &       & 0.7651 & 0.9133 & 0.8327 &       & 0.7744 & 0.9597 & 0.8571 &       & 0.7725 & 0.9674 & 0.8590 \\
    \midrule
          & A-O-S & 0.4681 & 0.5986 & 0.5254 &       & 0.2971 & 0.4085 & 0.3440 &       & 0.3508 & 0.5381 & 0.4247 &       & 0.4153 & 0.6109 & 0.4945 \\
          & A-O   & 0.4965 & 0.6348 & 0.5572 &       & 0.3454 & 0.4750 & 0.4000   &       & 0.3911 & 0.6000   & 0.4736 &       & 0.4437 & 0.6518 & 0.5280 \\
          & A-S   & 0.6295 & 0.8314 & 0.7165 &       & 0.4718 & 0.7235 & 0.5712 &       & 0.5262 & 0.8125 & 0.6388 &       & 0.5458 & 0.8429 & 0.6626 \\
          GPT-4o & O-S   & 0.5255 & 0.6824 & 0.5937 &       & 0.3890 & 0.5347 & 0.4504 &       & 0.4527 & 0.6513 & 0.5342 &       & 0.5102 & 0.6894 & 0.5864 \\
          CoT+few-shot & A     & 0.6795 & 0.8774 & 0.7658 &       & 0.5467 & 0.8337 & 0.6604 &       & 0.5898 & 0.8819 & 0.7069 &       & 0.5858 & 0.8916 & 0.7070 \\
          & O     & 0.5610 & 0.7342 & 0.6360 &       & 0.4395 & 0.6119 & 0.5116 &       & 0.5023 & 0.7215 & 0.5923 &       & 0.5465 & 0.7383 & 0.6281 \\
          & S     & 0.7927 & 0.9431 & 0.8614 &       & 0.7705 & 0.9220 & 0.8395 &       & 0.7890 & 0.9481 & 0.8613 &       & 0.7976 & 0.9703 & 0.8755 \\
   
    \bottomrule
    \hline
    \end{tabular}
}
    \caption{Performance of different types of element combinations in ABSA tasks using LLM.
}
\label{tab:gpt-results}
\end{table*}

\subsection{Case Study}
\label{sec:Case Study}

See Table \ref{tab:casestudy}.
\begin{table*}[ht]
\centering
\tiny
\renewcommand{\arraystretch}{1.5}
\scalebox{0.85}{
\begin{tabular}{p{2cm}@{\hskip 0.5cm}p{4cm}@{\hskip 0.5cm}p{4cm}@{\hskip 0.5cm}p{4cm}@{\hskip 0.5cm}p{0.8cm}@{\hskip 0.5cm}p{0.8cm}}
    \toprule
    
    Sentence & Ground Truth & \multicolumn{1}{l}{GPT results} & Ours  & Precision & Recall \\
    \midrule
    \midrule
    Creamy appetizers--taramasalata, eggplant salad, and Greek yogurt(with cuccumber, dill, and garlic) taste excellent when on warm pitas. & ['creamy appetizers', 'creamy', 'positive'],\newline ['creamy appetizers', 'excellent', 'positive'],\newline ['warm pitas', 'warm', 'neutral'],\newline ['taramasalata', 'creamy', 'positive'],\newline ['eggplant salad', 'excellent', 'positive'],\newline ['greek yogurt ( with cuccumber , dill , and garlic )', 'excellent','positive'] & ['appetizers', 'creamy', 'neutral'], \newline ['taramasalata', 'taste excellent', 'positive'], \newline ['eggplant salad', 'taste excellent', 'positive'], \newline ['greek yogurt', 'taste excellent', 'positive'], \newline ['warm pitas', 'warm', 'neutral'] & ['creamy appetizers', 'creamy', 'positive'],\newline ['creamy appetizers', 'excellent', 'positive'],\newline ['warm pitas', 'warm', 'neutral'],\newline ['eggplant salad', 'excellent', 'positive'],\newline ['greek yogurt , 'excellent','positive' & GPT: 1/5\newline Ours: 4/5 & GPT: 1/6\newline Ours: 4/6 \\
    \midrule
    We left without ever getting service. & ['service', 'without ever', 'negative'] & ['service', 'without ever getting', 'negative'] & ['service', 'without ever', 'negative'] & GPT: 0/1\newline Ours: 1/1 & GPT: 0/1\newline Ours: 1/1 \\
    \midrule
    I go out to eat and like my courses, servers are patient and never rush courses or force another drink.' & ['servers', 'patient', 'positive'] & ['courses', 'like', 'positive'], \newline ['servers', 'patient', 'positive'], \newline ['servers', 'never rush courses', 'positive'], \newline ['servers', 'never force another drink', 'positive'] & ['servers', 'patient', 'positive'],  & GPT: 1/5\newline Ours: 1/1 & GPT: 1/1\newline Ours: 1/1 \\
    \bottomrule
\end{tabular}
}
\caption{Case study}
\label{tab:casestudy}
\end{table*}

\end{document}